\pdfoutput=1

\documentclass[11pt]{article}

\usepackage[]{acl}

\usepackage{times}
\usepackage{latexsym}

\usepackage{graphics}
\usepackage{graphicx}
\usepackage{color}
\usepackage{colortbl}
\usepackage{amsfonts}
\usepackage{amssymb}
\usepackage{amsmath}
\usepackage{nicefrac}
\usepackage{bbm}
\usepackage{booktabs}
\usepackage{microtype}
\usepackage{multirow}
\usepackage{multicol}
\usepackage{tabu}
\usepackage{amsthm}
\usepackage{footmisc}
\usepackage{soul}
\usepackage{arydshln}
\usepackage{wasysym}
\usepackage{array}
\usepackage{makecell}

\usepackage{algpseudocode}
\usepackage{algorithm}
\usepackage{hyperref}
\usepackage{xspace}
\usepackage{url}
\usepackage{utfsym}
\usepackage{dirtytalk}

\usepackage{float}
\usepackage{enumitem}
\usepackage{tikz}
\usepackage{tcolorbox}
\usepackage{siunitx}
\usepackage{pifont}

\usepackage{CJKutf8}

\newcommand{\ours}{\textsc{Trice}}
\newcommand{\1}{\uppercase\expandafter{\romannumeral1}}
\newcommand{\2}{\uppercase\expandafter{\romannumeral2}}

\definecolor{aliceblue}{RGB}{178, 217, 245}
\newcommand{\CC}{\cellcolor{aliceblue}}
\definecolor{babyblue}{RGB}{217, 239, 251}
\newcommand{\CB}{\cellcolor{babyblue}}

\newcommand{\daugshifted}{\raisebox{0.5\depth}{$\uparrow$}}

\newcommand{\uaugshifted}{\raisebox{0.5\depth}{$\downarrow$}}

\newcommand{\uaulg}[1]{{\hlprimarytab{\uaugshifted{#1}}}}

\newcommand{\daulg}[1]{{\hlsecondarytab{\daugshifted{#1}}}}

\definecolor{c3}{cmyk}{0.9081,0,0.7209,0.5255}

\newtcbox{\hlprimarytab}{on line, rounded corners, box align=base, colback=c3!10,colframe=white,size=fbox,arc=3pt, before upper=\strut, top=-2pt, bottom=-4pt, left=-2pt, right=-2pt, boxrule=0pt}
\newtcbox{\hlsecondarytab}{on line, box align=base, colback=red!10,colframe=white,size=fbox,arc=3pt, before upper=\strut, top=-2pt, bottom=-4pt, left=-2pt, right=-2pt, boxrule=0pt}

\definecolor{my_green}{RGB}{51,102,0}
\definecolor{my_red}{RGB}{204, 0, 0}

\newcommand{\colorcmark}{\textcolor{my_green}{\ding{52}}}
\newcommand{\colorxmark}{\textcolor{my_red}{\ding{55}}}

\newcommand{\revise}[1]{\textcolor{black}{#1}}

\usepackage[T1]{fontenc}

\usepackage[utf8]{inputenc}

\usepackage{microtype}

\usepackage{inconsolata}

\title{Making Language Models Better Tool Learners with Execution Feedback}

\author{
    Shuofei Qiao\textsuperscript{1,2},
    Honghao Gui\textsuperscript{1,2},
    Chengfei Lv\textsuperscript{4}, \\
    \textbf{Qianghuai Jia\textsuperscript{4}},
    \textbf{Huajun Chen\textsuperscript{1,2,3}},
    \textbf{Ningyu Zhang\textsuperscript{1,2}}\thanks{$\quad$ Corresponding Author.} \\
 \textsuperscript{1}College of Computer Science and Technology, Zhejiang University \\
 \textsuperscript{2}ZJU-Ant Group Joint Research Center for Knowledge Graphs, Zhejiang University \\
 \textsuperscript{3}ZJU-Hangzhou Global Scientific and Technological Innovation Center, Zhejiang University
 \textsuperscript{4}Alibaba Group\\
    \texttt{
    \{shuofei,guihonghao,huajunsir,zhangningyu\}@zju.edu.cn 
    }
}

\begin{document}
\maketitle
\begin{abstract}
Tools serve as pivotal interfaces that enable humans to understand and reshape the environment. With the advent of foundation models, AI systems can utilize tools to expand their capabilities and interact with the real world. Existing tool learning methodologies, encompassing supervised fine-tuning and prompt engineering approaches, often induce large language models to utilize tools indiscriminately, as complex tasks often exceed their own competencies. However, introducing tools for simple tasks, which the models themselves can readily resolve, can inadvertently propagate errors rather than enhance performance. This leads to the research question: \emph{can we teach language models when and how to use tools?} To meet this need, we propose \textbf{T}ool lea\textbf{R}ning w\textbf{I}th exe\textbf{C}ution f\textbf{E}edback (\textbf{\ours}), a two-stage end-to-end framework that enables the model to continually learn through feedback derived from tool execution, thereby learning when and how to use tools effectively. Experimental results, backed by further analysis, show that {\ours} can make the large language model selectively use tools by improving the accuracy of tool usage while enhancing insufficient tool learning and mitigating excessive reliance on tools\footnote{Code: \url{https://github.com/zjunlp/TRICE}.}.
\end{abstract}

\section{Introduction}




The recent rapid advancement of foundation models~\citep{DBLP:conf/nips/BrownMRSKDNSSAA20,DBLP:conf/nips/Ouyang0JAWMZASR22,DBLP:journals/corr/abs-2204-02311,ACL2023_PromptReasoningSurvey,DBLP:journals/corr/abs-2303-18223} makes it practical for AI machines to create~\citep{DBLP:journals/corr/abs-2305-17126,DBLP:journals/corr/abs-2305-14318} and utilize tools effectively~\citep{DBLP:journals/corr/abs-2303-17580,DBLP:journals/corr/abs-2304-09842}, which greatly transcends their inherent limitations in various underlying areas, including arithmetic~\citep{DBLP:journals/corr/abs-2110-14168,DBLP:journals/corr/abs-2205-12255}, knowledge updating~\citep{DBLP:conf/iclr/Sun0TYZ23,DBLP:conf/acl/ZhaoLJQB23}, multi-modal semantic analysis~\citep{DBLP:journals/corr/abs-2303-04671,DBLP:journals/corr/abs-2303-03378}, etc.
\begin{figure}[t]
    \centering
    \resizebox{.48\textwidth}{!}{
    \includegraphics{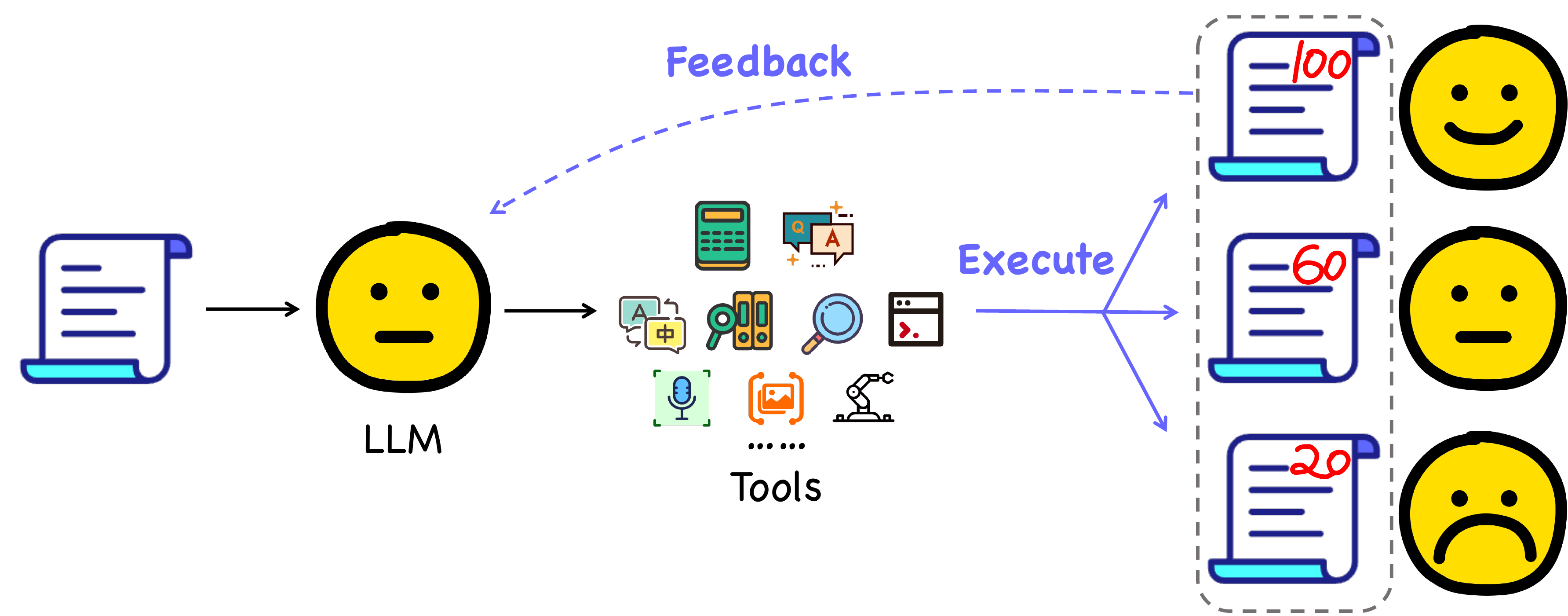}}
    \caption{Large language model learns to use tools from execution feedback.}
    \label{fig:intro}
\end{figure}
Existing research has shed light on the potential of Large Language Models (LLMs) to exhibit a promising level of dexterity and finesse in tool use~\citep{DBLP:journals/corr/abs-2304-08354,DBLP:journals/corr/abs-2305-13246}.
Prior works view tools as external resources to augment LLMs for better performance~\citep{DBLP:journals/corr/abs-2302-04761,DBLP:journals/corr/abs-2305-11554,DBLP:journals/corr/abs-2305-15334,DBLP:journals/corr/abs-2306-05301} or employ LLMs as a hub for human-tool interaction, responsible for orchestrating the deployment and usage of tools~\citep{DBLP:journals/corr/abs-2303-17580,DBLP:journals/corr/abs-2304-04370,DBLP:journals/corr/abs-2304-09842,DBLP:journals/corr/abs-2303-03378}.

\begin{table*}[t!]
\centering
\renewcommand\arraystretch{1}
\scalebox{0.7}{
\begin{tabular}{lccccccc}
\hline
\toprule
\textbf{Method} & \textbf{LM} & \textbf{Model Scale} & \textbf{Mechanism} & \textbf{Feedback} & \textbf{Peft} & \textbf{Teacher} & \textbf{Unseen} \\
\Xhline{1px}
Toolformer~\cite{DBLP:journals/corr/abs-2302-04761} & GPT-J & 6B & instruct-tuning & \colorxmark & \colorxmark & \colorxmark & \colorxmark \\
ToolkenGPT~\cite{DBLP:journals/corr/abs-2305-11554} & LLaMA & 13B, 30B & fine-tuning & \colorxmark & \colorxmark & \colorxmark & \colorxmark \\
HuggingGPT~\cite{DBLP:journals/corr/abs-2303-17580} & ChatGPT & >=100B & prompt & \colorxmark & \colorxmark & \colorxmark & \colorcmark \\
Chameleon~\cite{DBLP:journals/corr/abs-2304-09842} & GPT-4 & >=100B & prompt & \colorxmark & \colorxmark & \colorxmark & \colorcmark \\
ChatCoT~\cite{DBLP:journals/corr/abs-2305-14323} & ChatGPT & >=100B & prompt & \colorxmark & \colorxmark & \colorxmark & \colorcmark \\
Gorilla~\cite{DBLP:journals/corr/abs-2305-15334} & LLaMA & 7B & instruct-tuning & \colorxmark & \colorxmark & \colorcmark & \colorcmark \\
ToolLLaMA~\cite{DBLP:journals/corr/abs-2307-16789} & LLaMA & 7B & instruct-tuning & \colorxmark & \colorxmark & \colorcmark & \colorcmark \\
GPT4Tools~\cite{DBLP:journals/corr/abs-2305-18752} & Vicuna & 13B & instruct-tuning & \colorxmark & \colorcmark & \colorcmark & \colorcmark \\
\hline
{\ours} (ours) & \makecell{ChatGLM \\ Alpaca \\ Vicuna} & 6B, 7B & \makecell{instruct-tuning \\ reinforcement learning} & \colorcmark & \colorcmark & \colorcmark & \colorcmark \\
\bottomrule
\hline
\end{tabular}
}
\caption{Comparison of related works.  \textbf{Mechanism} denotes how the LM learns to invoke tools. \textbf{Feedback} stands for whether the LM learns from execution feedback. \textbf{Peft} means the parameter efficient tuning. \textbf{Teacher} expresses whether learning from a powerful teacher like ChatGPT. \textbf{Unseen} indicates the zero-shot capability on unseen tools.}
\label{tab:compare}
\end{table*}

Despite the empirical success of previous work, a critical issue remains: LLMs often do not understand \textbf{when and how to properly use which tools}.
On one hand, the use of tools is necessary to augment LLMs when facing complex problems that surpass their inherent capabilities. 
On the other hand, for simpler problems that can readily be solved by the models themselves, introducing tools can paradoxically propagate errors rather than enhance performance.
These errors can include but are not limited to, improper selection of tool types, generation of incorrect tool inputs, and ineffective utilization of tool return results.
Intuitively, it's crucial for LLMs to develop an awareness of when tools are necessary and when they are not, and to be able to make decisions about selecting the most appropriate tools for the task at hand.

To address the above issues, we propose \textbf{T}ool lea\textbf{R}ning w\textbf{I}th exe\textbf{C}ution f\textbf{E}edback (\textbf{\ours}) as shown in Figure~\ref{fig:intro}, a two-stage end-to-end framework that enables the model to continually learn through feedback derived from execution, thereby learning when and how to use tools effectively. 
Specifically, we first prepare a dataset that helps discern when tool usage is necessary for LLMs and when it is not.
Given the lack of gold labels, we utilize ChatGPT~\citep{ChatGPT-OpenAI} to automatically generate tool usage APIs.
Then, we introduce a two-stage training strategy to teach the model when to use tools:
1) \textbf{Behavior Cloning}. We conduct instruct-tuning on the dataset to let the model imitate the tool-using behavior.
2) \textbf{Reinforcement Learning with Execution Feedback (RLEF)}.
We further reinforce the model with execution feedback by aligning it with desirable candidate responses, guiding the model to selectively use tools to avoid error propagation.
We detail the main difference of {\ours} with related works in Table~\ref{tab:compare}.

We train and evaluate {\ours} on various tasks and backbone models.
Experimental results and further analyses demonstrate that {\ours} successfully instructs the model to judiciously use tools, simultaneously enhancing insufficient tool learning, reducing excessive reliance on tools, and improving the accuracy of tool usage. 
In summary, the key contributions of our study are as follows:
\begin{itemize}
\item We introduce {\ours}, a two-stage end-to-end training framework that leverages execution feedback to help LLMs become more proficient tool learners.
\item We perform superior on eight benchmark datasets of four tasks with various models.
\item Extensive empirical analysis demonstrates that {\ours} can guide the model in judicious tool use, thereby enhancing insufficient tool use, reducing excessive dependency on tools, and improving the effectiveness of tool use.
\end{itemize}

\section{Related Work}

\paragraph{Tool Learning.}
Though possessing remarkable capabilities~\citep{ACL2023_PromptReasoningSurvey,yao2023react}, LLMs still struggle in many basic aspects where much smaller and simpler tools may precisely excel.
Under this circumstance, a new paradigm, called Tool Learning~\citep{DBLP:journals/corr/abs-2304-08354}, is born to combine the strengths of both LLMs and specialized tools.
Some works~\citep{DBLP:journals/corr/abs-2303-03378,DBLP:journals/corr/abs-2303-17580,DBLP:journals/corr/abs-2304-09842} regard LLMs as a decision-making hub for compositional tool using which can be called Tool-Oriented Learning~\citep{DBLP:journals/corr/abs-2304-08354}, while others~\citep{DBLP:journals/corr/abs-2211-10435,liu2023minds,DBLP:journals/corr/abs-2302-04761} treat tools as complementary resources to extend the power of LLMs which can be called Tool-Augmented Learning~\citep{DBLP:journals/corr/abs-2302-07842,DBLP:journals/corr/abs-2304-08354}.
Despite their success, tool-augmented approaches tend to force LLMs to use tools mindlessly regardless of whether they actually need tools for help.
This may, in some scenarios, steer LMs to erroneously choose the type of tools or the way to use tools, making the loss outweighs the gain.
Compared to previous works, we try to \textbf{make LMs better tool learners by teaching them to use tools selectively instead of blindly}.

\begin{figure*}[t!]
    \centering
    \resizebox{.96\textwidth}{!}{
    \includegraphics{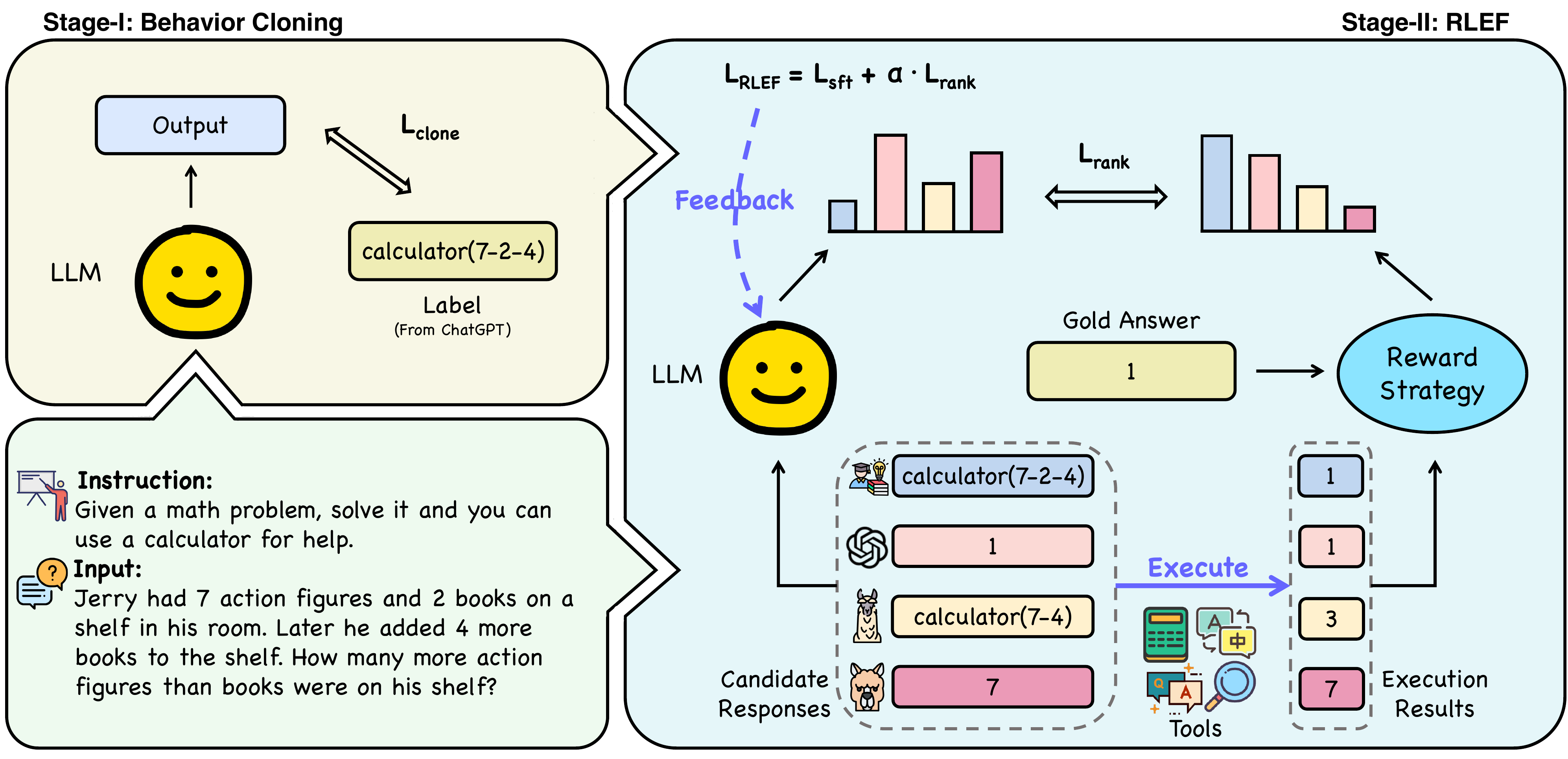}}
    \caption{The overview of our proposed framework \ours. In stage-\1 \textbf{(Behavior Cloning)}, We conduct instruct-tuning on the dataset to let the model imitate the tool-using behavior. 
    In stage-\2 \textbf{(RLEF)}, we further reinforce the model with tool execution feedback by aligning it with desirable candidate responses.}
    \label{fig:method}
\end{figure*}

\paragraph{Learning from Feedback.}
An intuitive approach of tool learning is to fit LLMs on examples with human-labeled tools directly~\citep{DBLP:conf/ijcai/TorabiWS18,DBLP:conf/nips/LiPPDWF0HAAAM0Z22}.
However, this is often impractical to annotate every possible scenario~\citep{DBLP:conf/iccv/CodevillaSLG19} and difficult to generalize to new ones.
It is worth noting that humans generally have the ability to correct and reflect on their own behavior from trial and error~\citep{DBLP:journals/corr/abs-1907-09620}.
Intuitively, feedbacks from the environments or humans enable LLMs to understand the impact of their actions and adapt their behavior accordingly.
Reinforcement learning (RL) excels at enabling models to learn decision-making abilities in complex environments through feedback~\citep{DBLP:journals/nature/SchrittwieserAH20,DBLP:conf/nips/Yao0YN22,DBLP:journals/corr/abs-2304-04370}.
\citet{DBLP:conf/nips/Ouyang0JAWMZASR22} apply a state-of-the-art RL algorithm, PPO~\citep{DBLP:journals/corr/SchulmanWDRK17}, to align LLMs with human feedback.
\citet{DBLP:conf/emnlp/0010HLHWHC22} reinforce knowledge for commonsense question answering with a fixed QA model providing feedback.
\revise{\citet{rrhf,rlaif} offer a promising alternative that leverages powerful off-the-shelf LLMs to generate preferences.}
Compared to the previous feedback framework, we introduce \textbf{RLEF} for tool learning which reinforces the LLMs with the execution result of tools.

\section{Methodology}
\label{method}

\paragraph{Problem Overview.}
We mainly focus on four kinds of tasks, with each instance in the format of $x = (s, q, t, a)$, where $s$ denotes the specialized instruction of each task, $q$ refers to the question, $t$ stands for the tool API and $a$ is the gold answer.
Following an instruction-following paradigm, the complete input of the LLM is as follows:
\begin{align}
\label{formula:input}
\mathtt{input} = [s, q],
\end{align}
where [,] stands for text concatenating.
In terms of the output, when LLM deems that no tool is necessary, it generates the answer $a$.
Conversely, if the model identifies the need for a tool, it produces the tool API $t$, which encompasses the specific type of tool and its corresponding input:

\begin{align}
\label{formula:output}
\mathtt{output=}
\begin{cases}
a &use\_tool=\mathtt{false} \\
t &use\_tool=\mathtt{true}
\end{cases}
\end{align}
The detailed format of each task is shown in \ref{app:task_format}.

Given the problem, the \textbf{main challenges} lie in 1) determining the LLM when to or not to harness tools for help and 2) how to impart the ability to the LLM to make selective use of tools.
For the \textbf{former}, we allow the untrained model to directly infer answers, considering the correct ones as not requiring tools and the incorrect ones as indicating the need for tool assistance.
For the \textbf{latter}, we adopt \textbf{\ours}, a two-stage training strategy.
In the first stage, we use \textbf{Behavior Cloning} to make the model imitate tools invoking.
Building upon this, we continue to train the model for selective tool usage with \textbf{RLEF} in the second stage. The overview of our method is illustrated in Figure~\ref{fig:method}.
\textbf{Please note that all symbols are globally defined in sections \ref{method}\&\ref{exp}.}

\paragraph{Data Preparation.}
The data preparation follows the principles outlined in Eq.\ref{formula:input}\&\ref{formula:output}.
Given a raw initial dataset $\mathcal{D}_\textrm{init}=\{(q, a)\}_{i=1}^{|\mathcal{D}_\textrm{init}|}$ from the benchmark, we utilize LLM without fine-tuning to generate predictions. Since we do not have gold labels for tool APIs, we employ ChatGPT~\citep{ChatGPT-OpenAI} to generate pseudo-labels under few-shot prompting. 
Specifically, we generate tool API labels $t=\mathtt{tool_{name}(tool_{input})}$ for questions where the generated predictions are incorrect.
For questions with correct predictions, we directly set $t=\mathtt{None}$ to indicate that tool APIs are not required. 
We design particular instructions $s$ tailored to each task.
In the end, we obtain $\mathcal{D}_\textrm{tool}=\{(s, q, t, a)\}_{i=1}^{|\mathcal{D}_\textrm{tool}|}$ according with Eq.\ref{formula:input}\&\ref{formula:output} containing the tool demand information of the specific LLM as we desire\footnote{For more details of data preparation, please refer to Appendix~\ref{app:data_construction}.}.

\paragraph{Training.} As shown in Figure~\ref{fig:method}, based on $\mathcal{D}_\textrm{tool}$, we conduct a two-stage training approach: 
\1) \textbf{Behavior Cloning} (\S \ref{sec:train1}). In this stage, we teach the model to imitate the tool usage behavior by fine-tuning it on $\mathcal{D}_\textrm{tool}$ in an instruct-tuning manner.
This empowers the model with preliminary functionality of tool API invocation.
\2) \textbf{Reinforcement Learning with Execution Feedback} (\S \ref{sec:train2}).
Drawing inspiration from fine-tuning with human feedback~\citep{DBLP:conf/nips/Ouyang0JAWMZASR22}, we continue to reinforce our model obtained in stage \1 with execution feedback by steering it to align with desirable candidate responses.

\subsection{Training Stage \1: Behavior Cloning}
\label{sec:train1}
During the behavior cloning stage, we aim to enable the LLM to master the schema of tool API invocation and develop preliminary skills in selectively utilizing tools.
We perform supervised fine-tuning on $\mathcal{D}_\textrm{tool}$ in this stage.

Specifically, for the model  $p_{\mathrm{LM}}$ with tunable parameters $\theta$, the training loss of stage \1 can be formulated as:
\begin{align}
\label{formula:ft}
    \mathcal{L}_\textrm{clone}(\theta) = \sum_{(s,q,t,a) \in \mathcal{D}_\textrm{tool}} - \log p_{\mathrm{LM}}(o|s,q;\theta),
\end{align}
where $o$ is the specified output of the model as defined in Eq.\ref{formula:output}.
The final parameterized model of this stage is denoted as $\theta_\textrm{clone}$.

\subsection{Training Stage \2: RLEF}
\label{sec:train2}
In stage \2, we continue to optimize $\theta_\textrm{clone}$ with execution feedback, so as to enhance its capability to selectively utilize tools and improve the accuracy of decision-making regarding tool types and corresponding inputs.

\paragraph{Overall Loss.}
Following \citet{rrhf}, for each question $q$, we have $k$ different candidate responses $y_i$ $(1 \leq i \leq k)$ marshaled from other LLMs (e.g. ChatGPT, LLaMA) or human experts.
We apply a reward strategy to score each $y_i$ with $r_i=R(a, y_i)$ where $a$ is the gold answer of question $q$.
Our goal is to instruct the LLM to determine the more desirable response by aligning with scores $\{r_i\}_k$.
So we then score each $y_i$ with the LLM:
\begin{align}
    p_i = \frac{\sum_t\log p_{\mathrm{LM}}(y_{i,m}|q,y_{i,<m};\theta_\textrm{clone})}{||y_i||},
\end{align}
where $m$ denotes the $m$th token of $y_i$, $p_i$ is the conditional log probability of $y_i$ and $||y_i||$ refers to the length-normalized factor.

To facilitate the LLM in learning the correct score ordering of different $y_i$, we introduce a ranking loss during training:
\begin{align}
\label{formula:rank}
    \mathcal{L}_\textrm{rank}=\sum_{r_i < r_j}\max(0,p_i - p_j).
\end{align}
Meanwhile, in order to prevent the model from deviating too far from the original parameters and generating unreasonable tool API invocation structure, we reintroduce the supervised fine-tuning loss:
\begin{align}
\label{formula:sft}
    \mathcal{L}_\textrm{sft}=-\sum_{m}\log p_{\mathrm{LM}}(o_m|s,q,o_{< m};\theta_\textrm{clone}).
\end{align}
Finally, the overall RLEF loss is defined as follows:
\begin{align}
\label{formula:policy}
    \mathcal{L}_\textrm{RLEF}=\alpha \cdot \mathcal{L}_\textrm{rank} + \mathcal{L}_\textrm{sft},
\end{align}
where $\alpha$ is a hyperparameter that determines the proportion of the rank loss.

\paragraph{Reward Strategy.}
The reward strategy aims to give each $y_i$ an $r_i$ and rank them accordingly for a given question $q$.
We view the output $o$ regulated in $\mathcal{D}_\textrm{tool}$ as the pseudo-human-expert (gold) response.
Then the reward strategy is derived from two indicators:
1) \textbf{accuracy of the answer} and
2) \textbf{consistency of tool usage with the gold response.}
Specifically, we employ a five-level scoring strategy.
We assign the gold response with the maximum score.
For the remaining, assuming that the correctness of the response is denoted as $\mathtt{True}$ for correct answers and $\mathtt{False}$ for incorrect answers, and whether the use of tools aligns with the gold response is denoted as $\mathtt{Yes}$ for alignment and $\mathtt{No}$ for misalignment, to ensure accurate and selective tool usage, our scoring is prioritized as follows:
$$
\mathtt{True}\mathtt{Yes} > \mathtt{True}\mathtt{No} > \mathtt{False}\mathtt{Yes} > \mathtt{False}\mathtt{No}.
$$
If two responses share the same state, they would receive the same score.

\begin{table}[t!]
\centering
\renewcommand\arraystretch{1}
\scalebox{0.63}{
\begin{tabular}{ccc}
\hline
\toprule
\textbf{Task} & \textbf{Tool} & \textbf{Datasets} \\
\hline
\textbf{\makecell{Math \\ Reasoning}} & Calculator & \makecell{ASDiv~\citep{DBLP:conf/acl/MiaoLS20} \\ SVAMP~\citep{DBLP:conf/naacl/PatelBG21} \\ GSM8K~\citep{DBLP:journals/corr/abs-2110-14168}} \\
\hline
\textbf{\makecell{Question \\ Answering}} & WikiSearch & \makecell{WebQuestions~\citep{DBLP:conf/emnlp/BerantCFL13} \\ NaturalQuestions~\citep{DBLP:journals/tacl/KwiatkowskiPRCP19} \\ TriviaQA~\citep{DBLP:conf/acl/JoshiCWZ17}} \\
\hline
\textbf{\makecell{LAMA}} & QA Model & T-REx~\citep{DBLP:conf/emnlp/PetroniRRLBWM19} \\
\hline
\textbf{\makecell{Multilingual \\ QA}} & Translator & MLQA~\citep{DBLP:conf/acl/LewisORRS20} \\
\bottomrule
\hline
\end{tabular}
}
\caption{Tasks, datasets and the corresponding tools.}
\label{tab:task}
\end{table}

\section{Experiments}
\label{exp}
\subsection{Experimental Settings}
\paragraph{Tasks and Tools.}
As shown in Table~\ref{tab:task}, we mainly evaluate our method on four tasks with each task specified to an external tool.
Due to limited computational resources, we randomly sample train and test sets from each dataset to reduce the data scale.
We display the detailed data distribution for each task in Figure~\ref{fig:dataset}.
Following \citet{DBLP:journals/corr/abs-2302-04761}, we use a more lenient evaluation criterion than exact match.
We simply check for the last number predicted by the model for the math reasoning task and check whether the correct answer is within the first twenty words for other tasks.
The QA model we use for LAMA is a retrieval-augmented LM fine-tuned on Natural Questions \citep{DBLP:journals/tacl/KwiatkowskiPRCP19} named \textit{Atlas} \citep{atlas}.
We use the 600M parameter NLLB \citep{DBLP:journals/corr/abs-2207-04672} as our machine Translation model for Multilingual QA.

\paragraph{Candidate Response Generation.}
We collect five responses for each question from four different models, e.g. ChatGPT, InstuctGPT, Vicuna-7B, Alpaca-7B, and the output regulated in $\mathcal{D}_\textrm{tool}$ as the gold response.
To differentiate whether or not to use tools among candidate responses, we compel ChatGPT and InstructGPT to utilize tools while allowing Alpaca and Vicuna to make the choice of using tools.
For ChatGPT and InstructGPT, we prompt them with instructions and few-shot examples, and for Alpaca-7B and Vicuna-7B, we fine-tune them on $\mathcal{D}_\textrm{tool}$ with LoRA~\citep{DBLP:conf/iclr/HuSWALWWC22} for a few steps in order to equip them with initial abilities for question answering and tool generation\footnote{For more details of candidate response generation, please refer to Appendix~\ref{app:response}.}.

\begin{table*}[t!]
\centering
\renewcommand\arraystretch{1.0}
\scalebox{0.67}{
\begin{tabular}{llccccccccc}
\hline
\toprule
{\multirow{2}{*}{\textbf{Setting}}}
& {\multirow{2}{*}{\textbf{Model}}} 
& \multicolumn{3}{c}{\textbf{Math Reasoning}} 
& \multicolumn{3}{c}{\textbf{Question Answering}}
& \multicolumn{1}{c}{\textbf{LAMA}}
& \multicolumn{1}{c}{\textbf{Multilingual QA}} 
& \multicolumn{1}{c}{\multirow{2}{*}{\textbf{Avg.}}} \\
\cmidrule(lr){3-5} \cmidrule(lr){6-8} \cmidrule(lr){9-9} \cmidrule(lr){10-10}
& & \textbf{ASDiv} & \textbf{SVAMP} & \textbf{GSM8K} & \textbf{WebQ} & \textbf{NaturalQ} & \textbf{TriviaQA} & \textbf{T-REx} & \textbf{MLQA} & \\
\Xhline{1px}
& GPT-3.5 
& 64.6 & 62.0 & 19.8
& \CC \textbf{46.4} & 15.0 & 41.3
& \CC \textbf{58.7}
& 34.4 
& 42.8 \\
\Xhline{1px}
& ChatGLM (Zero-Shot)
& 30.8 & 30.5 & 6.3
& 12.1 & 1.6 & 3.9
& 21.8
& 36.5
& 17.9 \\
\multirow{-2}{*}{\textit{\makecell{Prompt\\Based}}}
& ChatGLM (Few-Shot)
& 34.5 & 30.5 & 7.1
& 11.9 & 1.9 & 3.5
& 23.5
& 36.7
& 18.7 \\
\cmidrule(lr){1-2}
& ChatGLM (0\% Tool)
& 44.2 & 35.5 & 7.2
& 14.9 & 9.5 & 11.2
& 30.6
& 37.7
& 23.9 \\
\multirow{-2}{*}{\textit{\makecell{Supervised\\Fine-Tuning}}}
& ChatGLM (100\% Tool)
& 68.2 & 59.5 & 11.8
& 12.5 & 9.9 & 13.8
& 26.8
& 35.9
& 29.8 \\
\cmidrule(lr){1-2}
& \textbf{ChatGLM (\textsc{Trice-split})}
& 72.9 & \CB \underline{64.0} & 12.4
& 15.2 & 11.6 & 15.2
& 32.7
& 37.3
& 32.7 \\
\multirow{-2}{*}{\textit{\makecell{\textbf{\ours}\\\textbf{Based}}}}
& \textbf{ChatGLM (\textsc{Trice-mix})}
& \CB \underline{75.6} & \CC \textbf{65.5} & 15.8
& 18.5 & 13.7 & 29.0
& 34.7
& 41.7
& 36.8 \\
\Xhline{1px}
& Alpaca (Zero-Shot)
& 31.2 & 22.0 & 3.5
& 32.8 & 5.3 & 15.0
& 39.7
& 37.7
& 23.4 \\
\multirow{-2}{*}{\textit{\makecell{Prompt\\Based}}}
& Alpaca (Few-Shot)
& 38.3 & 23.5 & 4.3
& 33.9 & 6.0 & 16.6
& 41.1
& 45.5
& 26.2 \\
\cmidrule(lr){1-2}
& Alpaca (0\% Tool)
& 44.0 & 23.0 & 5.8
& 37.6 & 10.3 & 20.4
& 53.1
& 48.9
& 30.4 \\
\multirow{-2}{*}{\textit{\makecell{Supervised\\Fine-Tuning}}}
& Alpaca (100\% Tool)
& 68.6 & 44.5 & 15.6
& 35.9 & 16.4 & 32.6
& 41.7
& 46.6
& 37.7 \\
\cmidrule(lr){1-2}
& \textbf{Alpaca (\textsc{Trice-split})}
& 73.4 & 45.0 & 16.3
& 38.2 & 18.6 & 37.8
& 54.6
& 48.2
& 41.5 \\
\multirow{-2}{*}{\textit{\makecell{\textbf{\ours}\\\textbf{Based}}}}
& \textbf{Alpaca (\textsc{Trice-mix})}
& 75.2 & 58.0 & \CB \underline{21.5}
& 41.4 & \CB \underline{20.7} & \CB \underline{41.4}
& 55.2
& \CC \textbf{52.0}
& \CB \underline{45.7} \\
\Xhline{1px}
& Vicuna (Zero-Shot)
& 50.4 & 33.0 & 6.4
& 34.9 & 7.7 & 16.7
& 42.5
& 35.9
& 28.4 \\
\multirow{-2}{*}{\textit{\makecell{Prompt\\Based}}}
& Vicuna (Few-Shot)
& 56.1 & 35.5 & 6.9
& 36.2 & 8.8 & 17.6
& 44.2
& 38.5
& 30.5 \\
\cmidrule(lr){1-2}
& Vicuna (0\% Tool)
& 52.3 & 38.5 & 8.1
& 38.8 & 11.5 & 20.8
& 52.9
& 44.3
& 33.4 \\
\multirow{-2}{*}{\textit{\makecell{Supervised\\Fine-Tuning}}}
& Vicuna (100\% Tool)
& 69.4 & 48.0 & 15.8
& 37.1 & 17.5 & 33.9
& 45.7
& 42.1
& 38.7 \\
\cmidrule(lr){1-2}
& \textbf{Vicuna (\textsc{Trice-split})}
& 72.6 & 49.0 & 16.6
& 43.2 & \CB \underline{20.7} & 40.8
& 54.1
& 42.6
& 42.4 \\
\multirow{-2}{*}{\textit{\makecell{\textbf{\ours}\\\textbf{Based}}}}
& \textbf{Vicuna (\textsc{Trice-mix})}
& \CC \textbf{81.2} & 60.5 & \CC \textbf{21.8}
& \CB \underline{44.1} & \CC \textbf{21.2} & \CC \textbf{41.6}
&\CB \underline{55.4}
& \CB \underline{49.7}
& \CC \textbf{46.9} \\
\bottomrule
\hline
\end{tabular}
}
\caption{
Performance of {\ours} across various tasks with different backbone models.
\textbf{Zero-Shot:} models are directly evaluated.
\textbf{Few-Shot:} models are prompted by \revise{3} examples during evaluation.
\textbf{0\% Tool:} models are trained purely on question-answer paired data.
During the above settings, the model does not rely on tools.
\textbf{100\% Tool:} models are trained purely on question-tool paired data.
\textbf{\textsc{\ours-split}:} models are trained with \textsc{\ours} separately for each task.
\textbf{\textsc{\ours-mix}:} models are trained with \textsc{\ours} by combining training data from all tasks.}
\label{tab:main_result}
\end{table*}

\paragraph{Baselines.}
We mainly experiment with the following LLMs:
1) \textbf{GPT-3.5}~\citep{ChatGPT-OpenAI}. We utilize the \texttt{text-davinci-003} version of GPT series.
2) \textbf{ChatGLM-6B}~\citep{DBLP:conf/acl/DuQLDQY022}, a general language model pre-trained with an autoregressive blank-filling objective.
3) \textbf{Alpaca-7B}~\citep{alpaca}, a model further trained on LLaMA-7B~\citep{DBLP:journals/corr/abs-2302-13971} with self-instruction.
4) \textbf{Vicuna-7B}~\citep{vicuna2023}, an open-source chatbot trained by fine-tuning LLaMA-7B.
For GPT-3.5, we directly utilize the OpenAI API, while for other models, we train them all with LoRA~\citep{DBLP:conf/iclr/HuSWALWWC22} for efficiency in both stage-\1\&\2.

\begin{figure*}[t!]
    \centering
    \resizebox{0.985\textwidth}{!}{
    \includegraphics{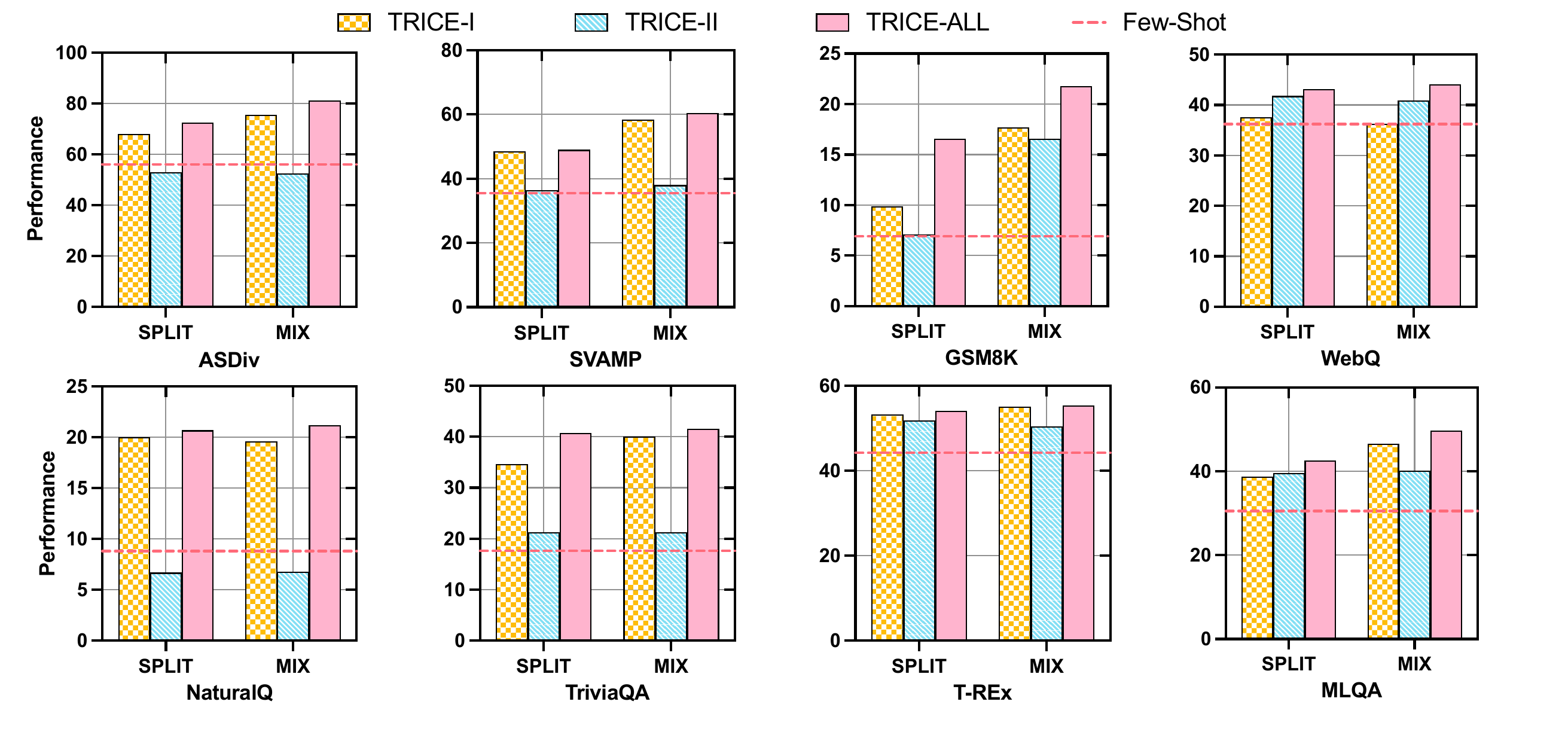}}
    \caption{Performance of {\ours} across all tasks at different training stages. \textbf{\textsc{\ours-\1}:} only train by Behavior Cloning (instruct-tuning) stage. \textbf{\textsc{\ours-\2}:} only train by RLEF (reinforcement learning with execution feedback) stage. \textbf{\textsc{\ours-all}:} train by both \textsc{\ours-\1} and \textsc{\ours-\2}.}
    \label{fig:ablation}
\end{figure*}

Based on the differences in training status, we classify the baselines of our primary experiment into three categories (see Table~\ref{tab:main_result}\&\ref{tab:generalization}):
1) \textit{Prompt-Based.} Models are directly evaluated without training under \textbf{Zero-Shot} or \textbf{Few-Shot} manners.
2) \textit{Supervised Fine-Tuning.} Models are trained purely on question-answer paired data (\textbf{0\% Tool} usage) or trained purely on question-tool paired data (\textbf{100\% Tool} usage).
3) \textit{\textsc{\ours}-Based.} Models are trained separately for each task (\textbf{\textsc{\ours-split}}) or by combining training data from all tasks (\textbf{\textsc{\ours-mix}}) with {\ours}.
Furthermore, we observe the role of each training stage (see Figure~\ref{fig:ablation}): 1) \textbf{\textsc{\ours-\1}.} Models are trained only by the Behavior Cloning stage. 2) \textbf{\textsc{\ours-\2}.} Models are trained only by the RLEF stage. 3) \textbf{\textsc{\ours-all}.} Models are trained by both \textsc{\ours-\1} and \textsc{\ours-\2}.
In our analysis, we use arrows to indicate \daulg{positive} and \uaulg{negative} performance compared to the specific baseline.

\paragraph{Setups.}
We fine-tune all our models with LoRA \cite{DBLP:conf/iclr/HuSWALWWC22} in the format proposed in Alpaca \cite{}.
All the models are trained for 5 epochs in stage-\1 and 2 epochs in stage-\2.
We use the learning rates of \{2e-5, 1e-4, 3e-4\} for ChatGLM-6B and \{2e-5, 1e-4\} for Alpaca-7B and Vicuna-7B.
The $\alpha$ is set to \{0.01, 0.1, 1\} for all the models.
The detailed hyper-parameters we use are shown in Appendix~\ref{app:training}.
Since sampling responses and training are separated, our whole training procedure only needs to load one model.
All our training can be completed on one 80G A800 GPU within 10 hours.

\subsection{Main Results}

\paragraph{Selective Tool Learning of Single Tool.}
Within each task, we train the model to learn the corresponding tool as shown in Table~\ref{tab:task}, thereby evaluating the model's proficiency in handling a single tool.
From the rows labeled \textsc{\ours-split} in Table~\ref{tab:main_result}, it is evident that training by {\ours}, Alpaca and Vicuna perform on par with GPT-3.5, exhibiting only a slight decrease of \uaulg{1.3\%} and \uaulg{0.4\%} on average.
Meanwhile, across all backbone models, \textsc{\ours-split} demonstrates significant improvements compared to the prompt-based baselines, surpassing the few-shot setting with \daulg{14.0\%} for ChatGLM, \daulg{15.3\%} for Alpaca, and \daulg{11.9\%} for Vicuna.
This indicates that {\ours} consistently empowers LLMs to use tools effectively, irrespective of the model architecture and scale (ChatGLM-6B is encoder-decoder, while Alpaca-7B and Vicuna-7B are decoder-only).
Moreover, whether it is completely independent (0\% Tool) or dependent (100\% Tool) on tools, supervised fine-tuning fails to beat {\ours}-based training, which highlights the necessity and efficacy of judicious tool learning.

\paragraph{Selective Tool Learning of Multi-Tools.}
Across all tasks, we train the model to simultaneously learn all the tools, assessing its capabilities in multi-tool learning.
As indicated in the rows labeled \textsc{\ours-mix} in Table~\ref{tab:main_result}, training across tasks achieves state-of-the-art performance by further exceeding the \textsc{\ours-split} with over \daulg{4.0\%} average score gains across different models.
Meanwhile, both Alpaca and Vicuna outperform GPT-3.5, exhibiting improvements of \daulg{2.9\%} and \daulg{4.1\%}, respectively.
These results declare the potential of {\ours} in selective multi-tool learning, which paves the way for expanding the capabilities of LLMs to wisely handle more complex and diverse types of tools.

\begin{table}[t!]
\centering
\renewcommand\arraystretch{1.0}
\scalebox{0.58}{
\begin{tabular}{lcccc}
\hline
\toprule
\multirow{3}{*}{\textbf{Model}} & \multicolumn{3}{c}{\textbf{Unseen Dataset}} & \multicolumn{1}{c}{\textbf{Unseen Tool}} \\
\cmidrule(lr){2-4} \cmidrule(lr){5-5}
& \multicolumn{2}{c}{\textbf{Calculator}} & \textbf{QA Model} & \textbf{Retriever} \\
\cmidrule(lr){2-3} \cmidrule(lr){4-4} \cmidrule(lr){5-5}
& \textbf{MultiArith} & \textbf{AddSub} & \textbf{SQuAD} & \textbf{HotpotQA} \\
\Xhline{1px}
GPT-3.5 & 51.1 & 59.5 & \CC \textbf{45.2} & \CC \textbf{36.7} \\
\Xhline{1px}
Vicuna (Zero-Shot) & 42.3 & 44.1 & 28.6 & 19.7 \\
Vicuna (Few-Shot) & 45.5 & 49.1 & 31.2 & 20.6 \\
\textbf{Vicuna (\textsc{\ours-split})} & \CB \underline{63.1} & \CB \underline{75.2} & 30.9 & --- \\
\textbf{Vicuna (\textsc{\ours-mix})} & \CC \textbf{66.6} & \CC \textbf{80.5} & \CB \underline{35.7} & \CB \underline{27.3} \\
\bottomrule
\hline
\end{tabular}
}
\caption{Performance to unseen datasets and tools. The presence of an empty value in the Unseen Tool TRICE-APLIT section is due to our reliance on the generalization achieved through mixed-tool training (MIX). Testing the new tool individually through separate training (SPLIT) using any single tool would not be suitable.}
\label{tab:generalization}
\end{table}

\begin{figure*}[t!]
    \centering
    \resizebox{0.985\textwidth}{!}{
    \includegraphics{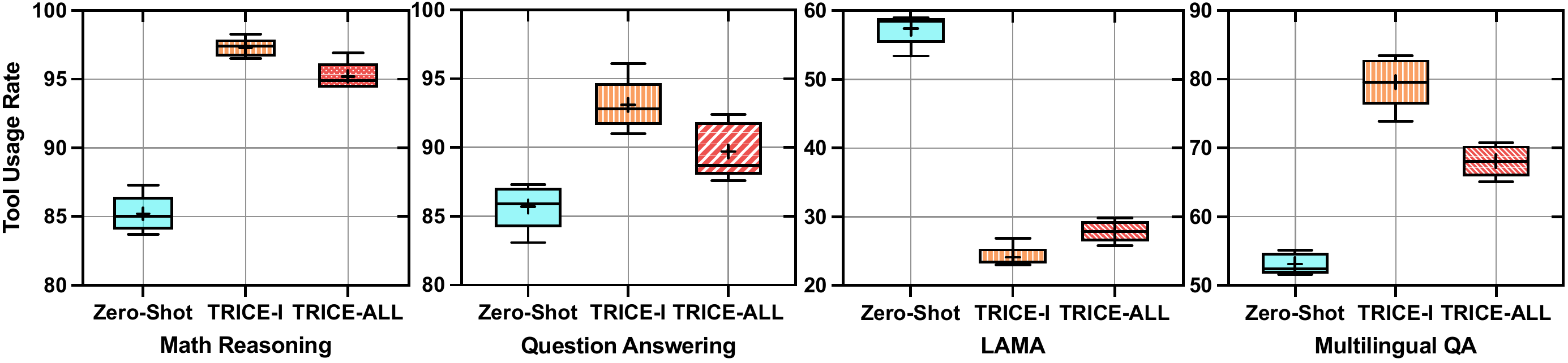}}
    \caption{Comparison of tool use rate statistics among different training stages. 
    In the Zero-Shot stage, we consider a need for tools when the model reaches a wrong answer.}
    \label{fig:tool_using_rates}
\end{figure*}

\paragraph{Generalization of Tool Learning.}
To analyze the generalization ability of {\ours}, we extend the trained model to unseen datasets and tools.
As illustrated in Table~\ref{tab:generalization}, we evaluate Vicuna on another two math reasoning datasets (MultiArith~\citep{DBLP:conf/emnlp/RoyR15} and AddSub~\citep{DBLP:conf/emnlp/HosseiniHEK14}) and one LAMA dataset (SQuAD~\citep{DBLP:conf/emnlp/PetroniRRLBWM19}).
Our approach enables continuous optimization of the model's performance on unseen datasets, with \textsc{\ours-mix} yielding superior results compared to \textsc{\ours-split}.
This suggests that {\ours} equips the model with general tool usage capabilities.
Furthermore, we steer the model towards unseen tools by simply modifying the instructions.
The performance of Vicuna (\textsc{\ours-mix}) augmented by a retriever on HotpotQA~\citep{DBLP:conf/emnlp/Yang0ZBCSM18} advances \daulg{6.7\%} than the few-shot manner.
Despite the disparities between GPT-3.5 on certain datasets, these findings highlight the promise of multi-tool training based on {\ours} for facilitating the generalization of tool learning.

\subsection{Analyses of Selective Tool Learning}
\paragraph{Stage-\1 is the Foundation of Stable Selective Tool Learning.}
Figure~\ref{fig:ablation} showcases the performance of {\ours} at different training stages, with Vicuna as the representative.
It is evident that only trained in stage \1 (\textsc{Trice-\1}), the model acquires efficacious tool usage capabilities, resulting in a substantial performance improvement.
Upon further training in stage \2 (\textsc{Trice-all}), the model experiences additional performance enhancements in both the \textsc{Split} and \textsc{Mix} training settings.
However, the results obtained solely from stage \2 (\textsc{Trice-\2}) are unsatisfactory, indicating that the initial tool generation ability bestowed upon the model during stage \1 is crucial for more stable training.

\begin{figure*}[t!]
    \centering
    \resizebox{.9\textwidth}{!}{
    \includegraphics{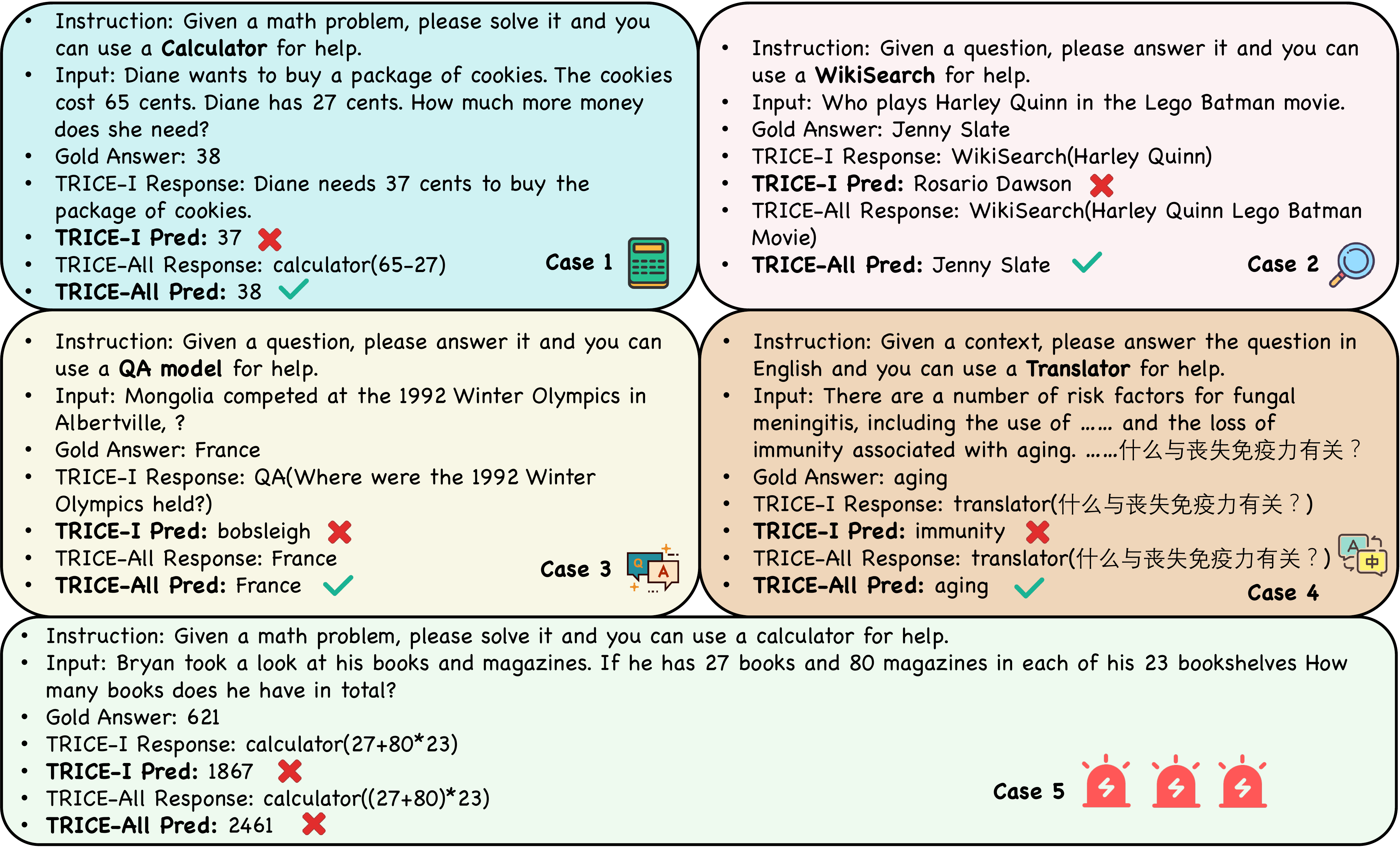}}
    \caption{Case study.
    We mainly show the responses and predictions of stages \1 and All.}
    \label{fig:case_study}
\end{figure*}

\paragraph{Stage-\2 Plays a Pivotal Role in Selective Tool Learning.}
To investigate how the models learn to use tools selectively, we analyze the tool usage rate statistics of Vicuna during each training stage in Figure~\ref{fig:tool_using_rates}.
After stage \1, we notice that the model's reliance on tools has significantly deepened on most tasks.
This indicates that the model effectively learns the pattern of tool usage in stage \1. Still, due to the imbalanced data distribution regarding the presence or absence of tools in the training set, instruct-tuning tends to make the model overly dependent on tools.
However, after stage \2, the model not only shows performance improvement (see Figure~\ref{fig:ablation}) but visibly reduces its dependency on tools, which illustrates that the execution feedback can help mitigate the model's excessive reliance on tools and alleviate error propagation in tool usage.
Moreover, it cannot be ignored that the fluctuation of LAMA differs from others.
The decision-making process for invoking the QA model poses challenges, leading to insufficient tool learning during stage \1.
The improvement in tool usage rate during stage \2 implies that the execution feedback can help address the issue of inadequate tool learning.
The above two phenomena highlight the validity of {\ours} for selective tool usage.

\paragraph{Case Study.}
In Figure~\ref{fig:case_study}, we present several cases featuring responses and predictions from different stages.
Case 1 suggests that stage \2 can alleviate the insufficient tool learning in stage \1, urging the model to seek assistance from tools for questions it struggles to answer.
Though stage \1 equips the model with a certain level of tool generation capability, it may not excel in making optimal decisions about the tool's input, as shown in Case 2.
Stage \2 mitigates this limitation and enhances the accuracy of tool use.
Case 3 confirms that our proposed method enables the model to use tools judiciously.
In Case 4, despite having the same tool invocation in both stages \1 and \2, the model may generate completely opposite answers. This indicates that stage \2 can further optimize the model's ability to leverage the return results of tools.
However, as shown in Case 5, our model still exhibits certain flaws leading to errors in tool usage.
We speculate that this could be attributed to the scale of our backbone models, which generally range from 6-7B, potentially limiting their tool learning ability.

\section{Discussion}
\paragraph{Knowledge Conflicts.}
In tool learning, LLMs manipulate tools and respond to users conditioned on a variety of knowledge sources.
One particularly challenging issue in tool learning is the problem of knowledge conflicts~\citep{DBLP:journals/corr/abs-2304-08354} which may derive from the conflicts between model knowledge and augmented knowledge from tools, and among augmented knowledge from different tools.
This may lead to a lack of explainability in model prediction and planning.
LLMs need to have the ability to differentiate knowledge from various sources and discern which ones are valuable, which ones are irrelevant, and even which ones may be harmful.
This ability becomes even more critical in highly specialized fields such as biomedical research and legal assistance, where the accuracy and reliability of knowledge are of utmost importance.
Our approach leverages the feedback loop of trial and error (see Figure~\ref{fig:method}) to learn when to use tools and when not to.
The model learns to recognize situations where relying solely on its intrinsic knowledge may not be sufficient and utilizing tools is more reliable.
Similarly, it learns to identify scenarios where its own learned knowledge is capable of solving the problem without the need for extensive tool usage. 
This learning process allows the model to adapt and make informed decisions about when to rely on its own capabilities and when to utilize tools effectively (see Figure~\ref{fig:tool_using_rates}).

\paragraph{Interactive Learning.}
Recent NLP has witnessed rapid advancement in interactive learning which considers language models as agents capable of observing, acting, and receiving feedback in a loop with external objects such as humans, knowledge bases, tools, models, and environments \citep{DBLP:journals/corr/abs-2305-13246}.
Collaboration among multi-agents \citep{DBLP:journals/corr/abs-2305-17390,DBLP:journals/corr/abs-2305-19118} and learning from feedback \citep{DBLP:journals/corr/abs-2209-09874,DBLP:conf/corl/IchterBCFHHHIIJ22} are the keys to achieving general embodied intelligence as of now.
Our approach is a preliminary endeavor to explore the incorporation of embodied methods into tool learning.
By leveraging feedback obtained through interactions between the environment (tools) and multi-agents with varying capabilities, we enable language models to learn more desirable execution strategies (see Figure~\ref{fig:case_study}).
However, our current method is unable to learn the usage of multi-tool compositions.
In the future, more sophisticated trial-and-error processes and feedback mechanisms will be necessary for LLMs to better utilize tools (e.g. learning multi-tool compositions) and even create new tools.

\section{Conclusion}

In this paper, we focus on addressing the challenge of selective utilization of tools by LLMs and propose a two-stage end-to-end training framework dubbed {\ours} to make LLMs better tool learners with execution feedback.
Through comprehensive experiments, we show that our method can achieve better performance compared to GPT-3.5.
Further analyses illustrate that {\ours} can selectively use tools by improving the accuracy of tool usage while enhancing insufficient tool learning and mitigating excessive reliance on tools.


\section*{Limitations}


In this paper, we focus on addressing the challenge of selective utilization of external tools by LLMs and propose a two-stage end-to-end training framework dubbed {\ours} to make LLMs better tool learners with execution feedback.
Despite our best efforts, there may be still some limitations that remain in this paper.

\paragraph{Method.}
Our approach can be applied to any tool-learning scenario, including embodied robotics.
However, due to the iterative nature of execution feedback, which relies on continuous trial-and-error, it is typically more suitable for computationally feasible virtual environments, while real-world scenarios often require a significant time investment.
In the future, we will explore more scientific and intricate feedback mechanisms to address the limitations above.

\paragraph{Language Models.}
Given our limited computational resources, we only conduct experiments on three backbone models with scales of 6-7B.
In the future, we may advent on LLMs with different architectures and larger scales.

\paragraph{Tasks and Datasets.}
Due to the limited resources, we only experiment on four tasks containing eight datasets.
There are also numerous tasks and scenarios that require the utilization of more diverse and complex tools.
In the future, we will embark on further research endeavors.

\section*{Acknowledgements}
We would like to express gratitude to the anonymous reviewers for their kind comments.
This work was supported by the National Natural Science Foundation of China (No. 62206246), the Fundamental Research Funds for the Central Universities (226-2023-00138), Zhejiang Provincial Natural Science Foundation of China (No. LGG22F030011), CAAI-Huawei MindSpore Open Fund, Ningbo Natural Science Foundation (2021J190), Yongjiang Talent Introduction Programme (2021A-156-G), CCF-Tencent Rhino-Bird Open Research Fund, and Information Technology Center and State Key Lab of CAD\&CG, Zhejiang University.

\bibliography{anthology,custom}

\appendix
\section{Appendix}
\label{sec:appendix}

\begin{table}[t!]
\centering
\renewcommand\arraystretch{1}
\scalebox{0.7}{
\begin{tabular}{ccc}
\hline
\toprule
\textbf{Name} & \textbf{Stage-\1} & \textbf{Stage-\2} \\
\hline
lora\_r & 8 & 8 \\
lora\_alpha & 16 & 16 \\
lora\_target\_modules & q\_proj v\_proj & q\_proj v\_proj \\
lora\_dropout & 0.05 & 0.05 \\
max\_length & 2048 & 2048 \\
batch\_size\_per\_device & 48 & 8 \\
gradient\_accumulation\_steps & 8 & 32 \\
warmup\_steps & 0 & 0 \\
epochs & 5 & 2 \\
lr & 1e-4, 3e-4 & 1e-4, 2e-5 \\
$\alpha$ & --- & 0.01, 0.1, 1 \\
\bottomrule
\hline
\end{tabular}
}
\caption{Hyperparameters to train Chatglm-6B.}
\label{tab:chatglm_para}
\end{table}

\begin{table}[t!]
\centering
\renewcommand\arraystretch{1}
\scalebox{0.7}{
\begin{tabular}{ccc}
\hline
\toprule
\textbf{Name} & \textbf{Stage-\1} & \textbf{Stage-\2} \\
\hline
lora\_r & 8 & 8 \\
lora\_alpha & 16 & 16 \\
lora\_target\_modules & q\_proj v\_proj & q\_proj v\_proj \\
lora\_dropout & 0.05 & 0.05 \\
max\_length & 512 & 512 \\
batch\_size\_per\_device & 128 & 8 \\
gradient\_accumulation\_steps & 8 & 32 \\
warmup\_steps & 0 & 0 \\
epochs & 5 & 2 \\
lr & 1e-4, 2e-5 & 1e-4, 2e-5 \\
$\alpha$ & --- & 0.01, 0.1, 1 \\
\bottomrule
\hline
\end{tabular}
}
\caption{Hyperparameters to train Alpaca-7B and Vicuna-7B.}
\label{tab:alpaca_para}
\end{table}

\begin{figure}[t!]
    \centering
    \resizebox{.48\textwidth}{!}{
    \includegraphics{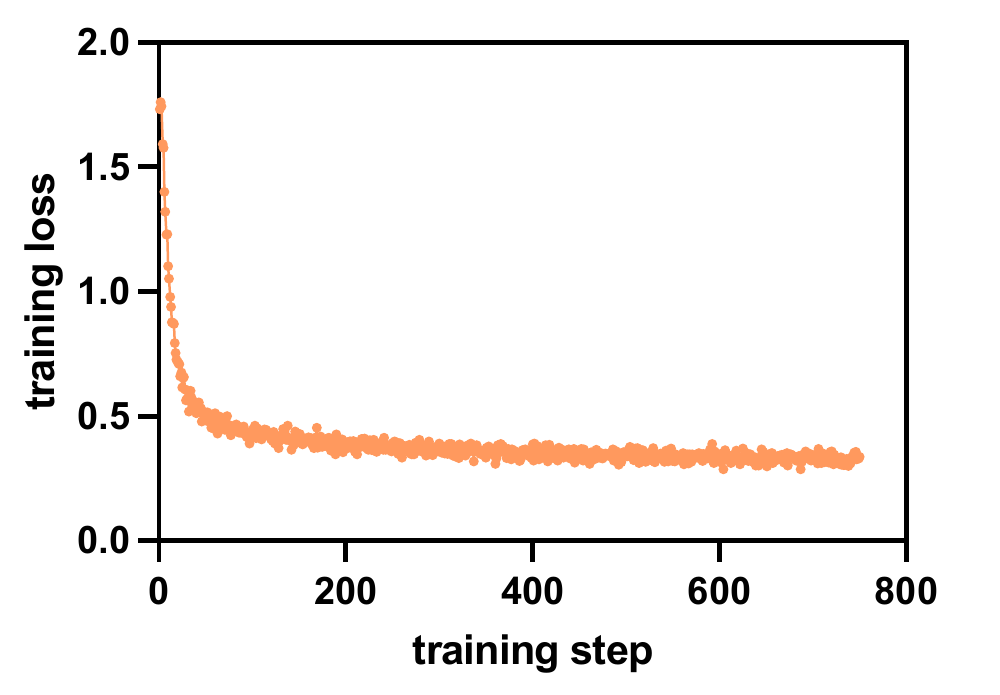}}
    \caption{Training loss variations of Vicuna-7B in stage \1 of \textsc{\ours-mix}.}
    \label{fig:sft_loss}
\end{figure}

\begin{figure}[t!]
    \centering
    \resizebox{.48\textwidth}{!}{
    \includegraphics{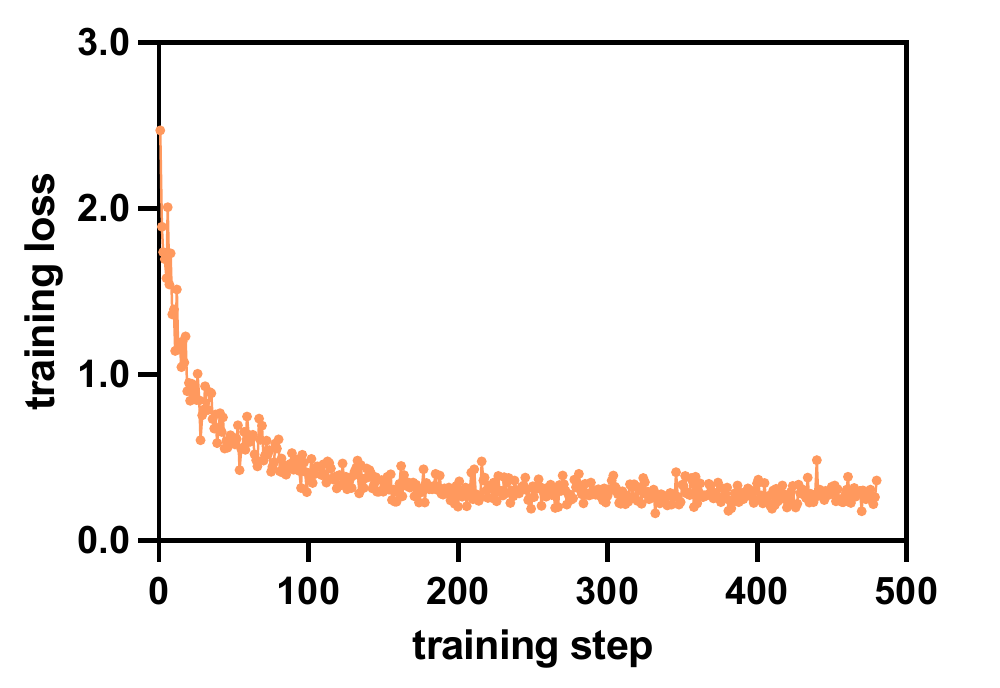}}
    \caption{Training loss variations of Vicuna-7B in stage \2 of \textsc{\ours-mix}.}
    \label{fig:rl_loss}
\end{figure}

\subsection{Task Format}
\label{app:task_format}
We mainly evaluate our method on four kinds of tasks as shown in Table~\ref{tab:task}.
Eq.\ref{formula:input}\&\ref{formula:output} formally define the input and output of each task in general.
Here is the detailed format of each task.

\paragraph{Math Reasoning}
\textbf{:}

Instruction $s$: \texttt{Given a math problem, please solve it and you can use a calculator for help.}

Question $q$: \texttt{Mrs. Hilt has 50 cents. A pencil costs 5 cents. How many pencils can she buy with the money she has?}

Tool API $t$ (if needed): \texttt{calculator(50/5)}

Gold Answer $a$: \texttt{10}

\paragraph{Question Answering}
\textbf{:}

Instruction $s$: \texttt{Given a question, please answer it and you can use a WikiSearch for help.}

Question $q$: \texttt{Where are sunbeam microwaves made?}

Tool API $t$ (if needed): \texttt{WikiSearch(Sunbeam microwaves manufacturing location)}

Gold Answer $a$: \texttt{Florida}

\paragraph{LAMA}
\textbf{:}

Instruction $s$: \texttt{Given a question, please answer it and you can use a QA model for help.}

Question $q$: \texttt{Winners of the festivals «Chervona Ruta» (Ukraine), «Pearls of the Season» (Ukraine), «Boards»
(Moscow), «Woodstock» ( ?}

Tool API $t$ (if needed): \texttt{QA(Where is the Woodstock festival held?)}

Gold Answer $a$: \texttt{Poland}

\paragraph{Multilingual QA}
\begin{CJK}{UTF8}{gbsn}
\textbf{:}

Instruction $s$: \texttt{Given a context, please answer the question in English and you can use a translator for help.}

Question $q$: \texttt{Context: Over the next decade, she went on more than 40 field missions, meeting with refugees and
internally displaced persons in over 30 countries. In 2002, when asked what she hoped to accomplish, she
stated, “Awareness of the plight of these people. I think they should be commended for what they have
survived, not looked down upon.” To that end, her 2001–02 field visits were chronicled in her book Notes
from My Travels, which was published in October 2003 in conjunction with the release of her
humanitarian drama Beyond Borders.
Question: 她在10年内完成了多少任务?}

Tool API $t$ (if needed): \texttt{translator(她在10年内完成了多少任务?)}

Gold Answer $a$: \texttt{more than 40}
\end{CJK}

\subsection{Data Preparation}
\label{app:data_construction}
\begin{figure*}[t!]
    \centering
    \resizebox{.96\textwidth}{!}{
    \includegraphics{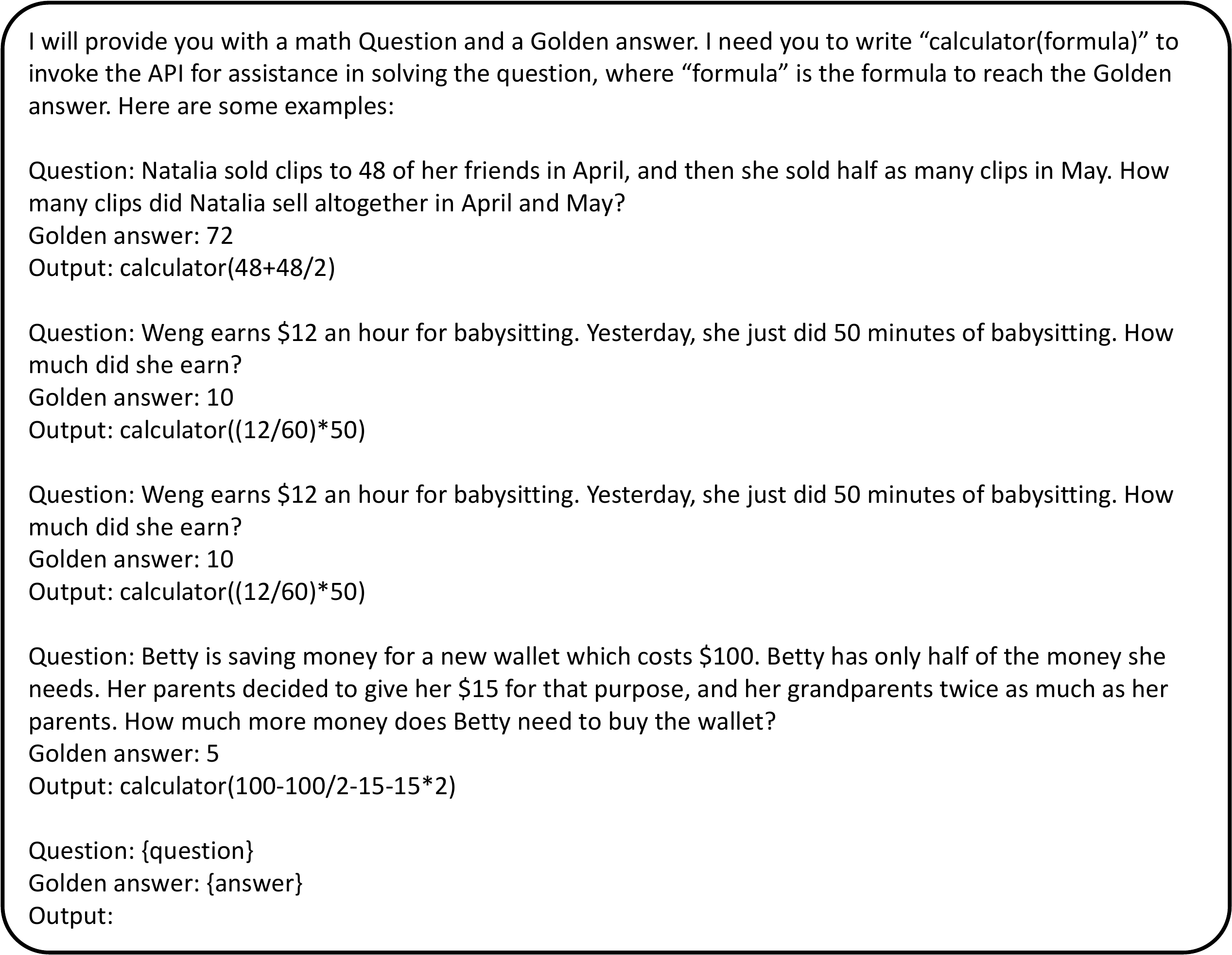}}
    \caption{Prompt used for Math Reasoning to generate tool APIs.}
    \label{fig:math_prompt_1}
\end{figure*}

\begin{figure*}[t!]
    \centering
    \resizebox{.96\textwidth}{!}{
    \includegraphics{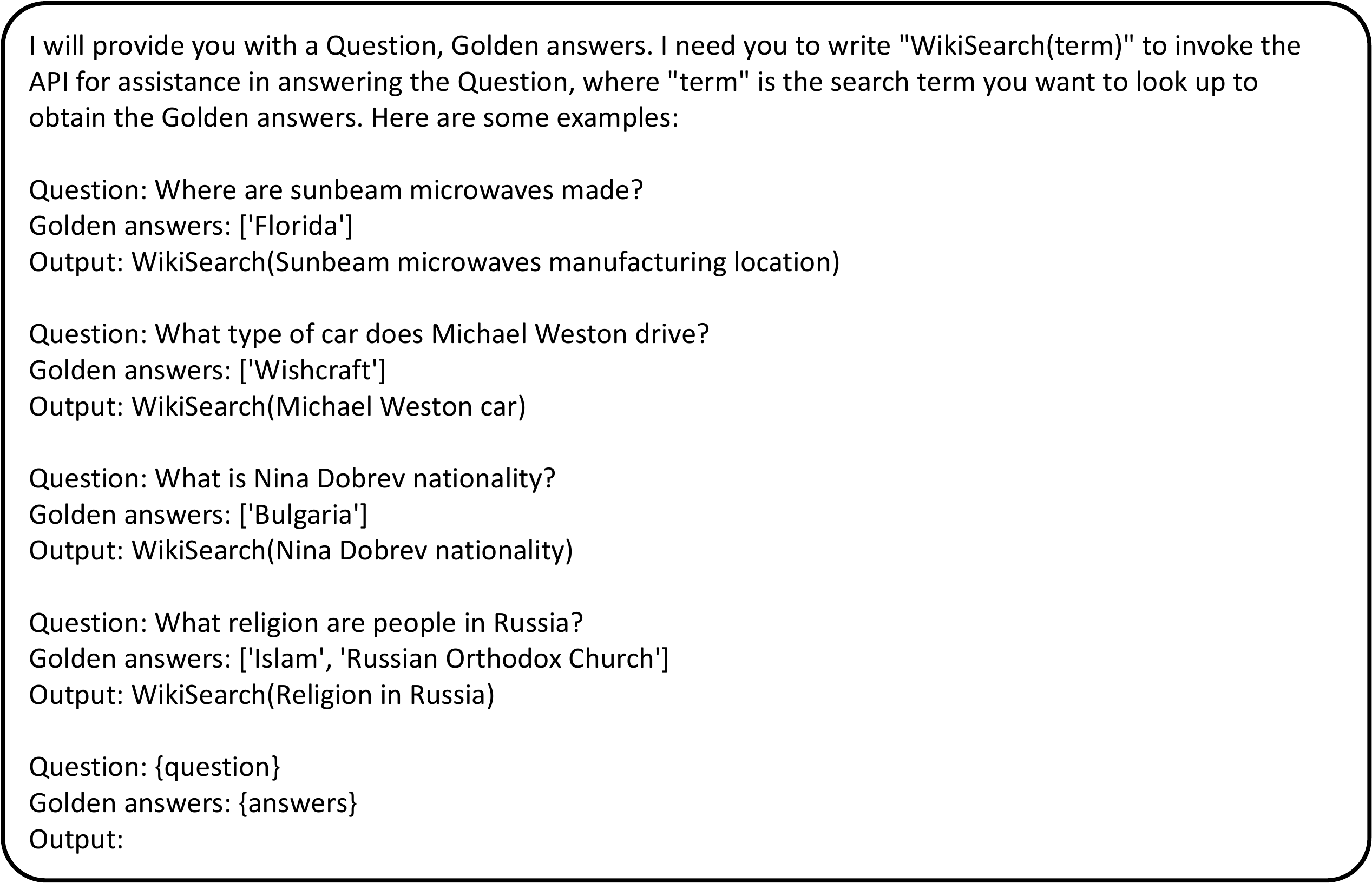}}
    \caption{Prompt used for Question Answering to generate tool APIs.}
    \label{fig:qa_prompt_1}
\end{figure*}

\begin{figure*}[t!]
    \centering
    \resizebox{.96\textwidth}{!}{
    \includegraphics{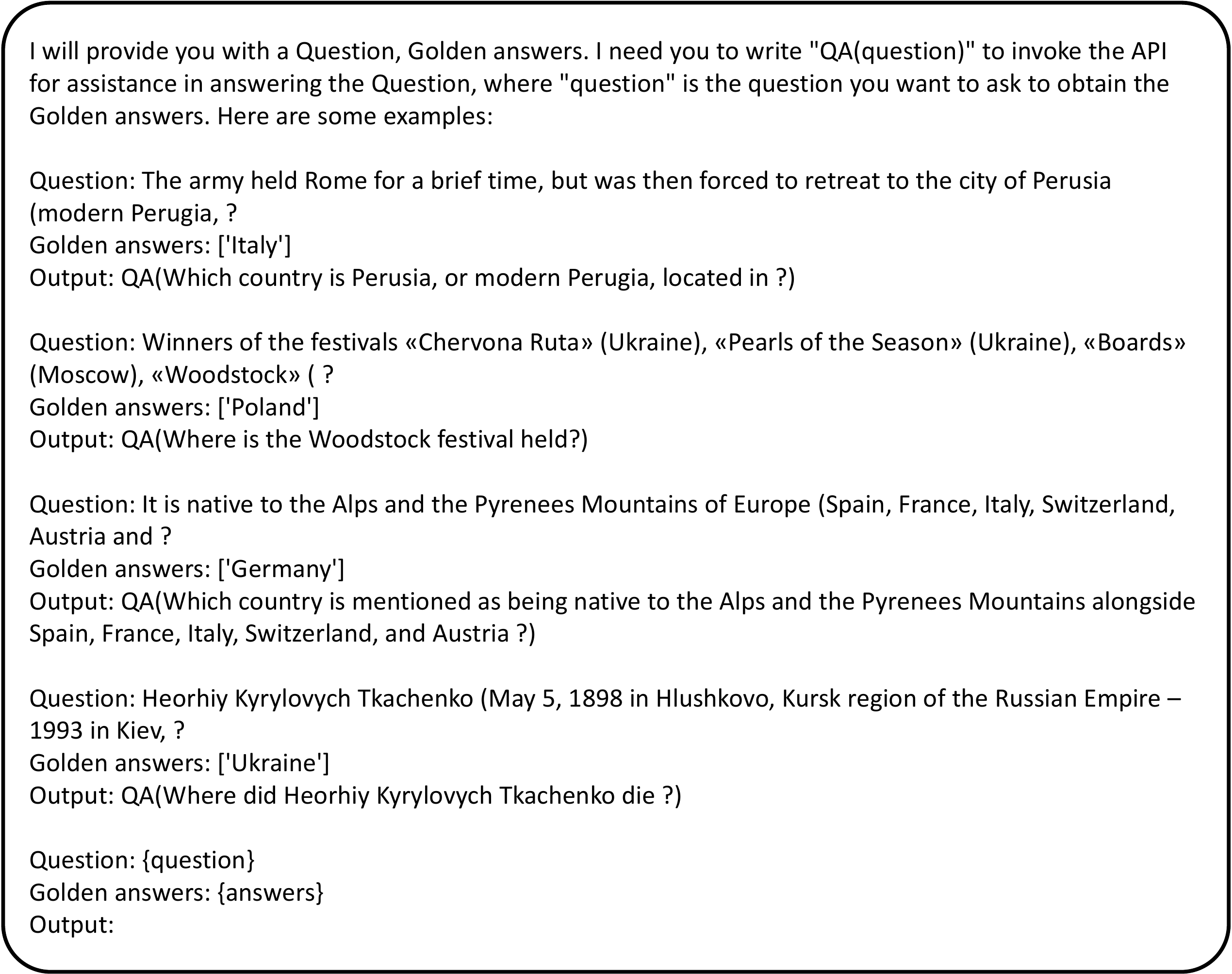}}
    \caption{Prompt used for LAMA to generate tool APIs.}
    \label{fig:lama_prompt_1}
\end{figure*}

\begin{figure*}[t!]
    \centering
    \resizebox{.96\textwidth}{!}{
    \includegraphics{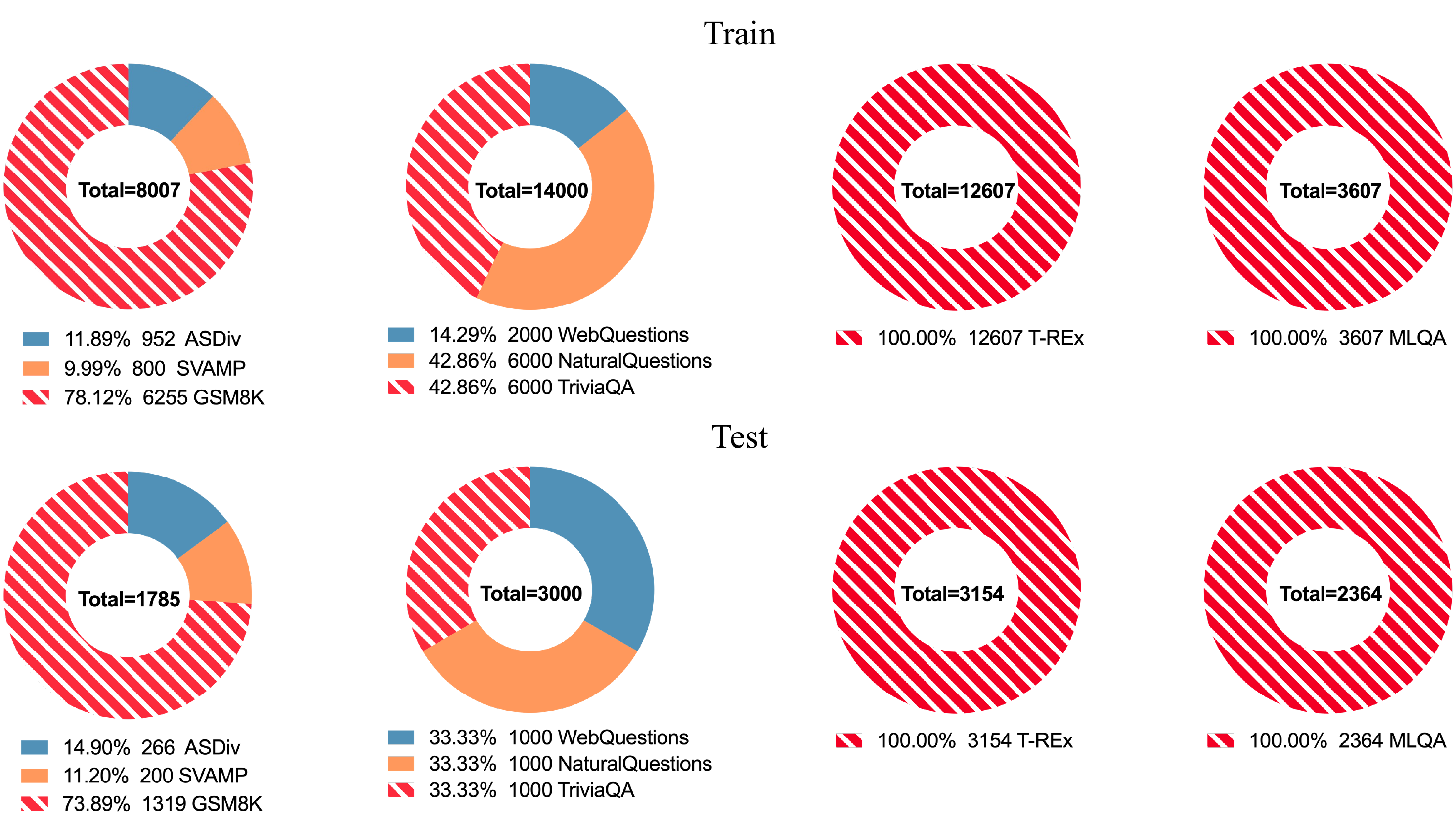}}
    \caption{Data distribution for each task.}
    \label{fig:dataset}
\end{figure*}

We present the prompt used to generate tool APIs for Math Reasoning, Question Answering, and LAMA in Figure~\ref{fig:math_prompt_1}-\ref{fig:lama_prompt_1}.
Since the sentence to be translated happens to be the provided question, Multilingual QA does not require ChatGPT to generate tool APIs.
Due to limited computational resources, we randomly sample train and test sets from each dataset to reduce data scale and training/testing costs.
The final data distribution for each task is illustrated in Figure~\ref{fig:dataset}.

\subsection{Response Generation}
\label{app:response}
\begin{figure*}[t!]
    \centering
    \resizebox{.96\textwidth}{!}{
    \includegraphics{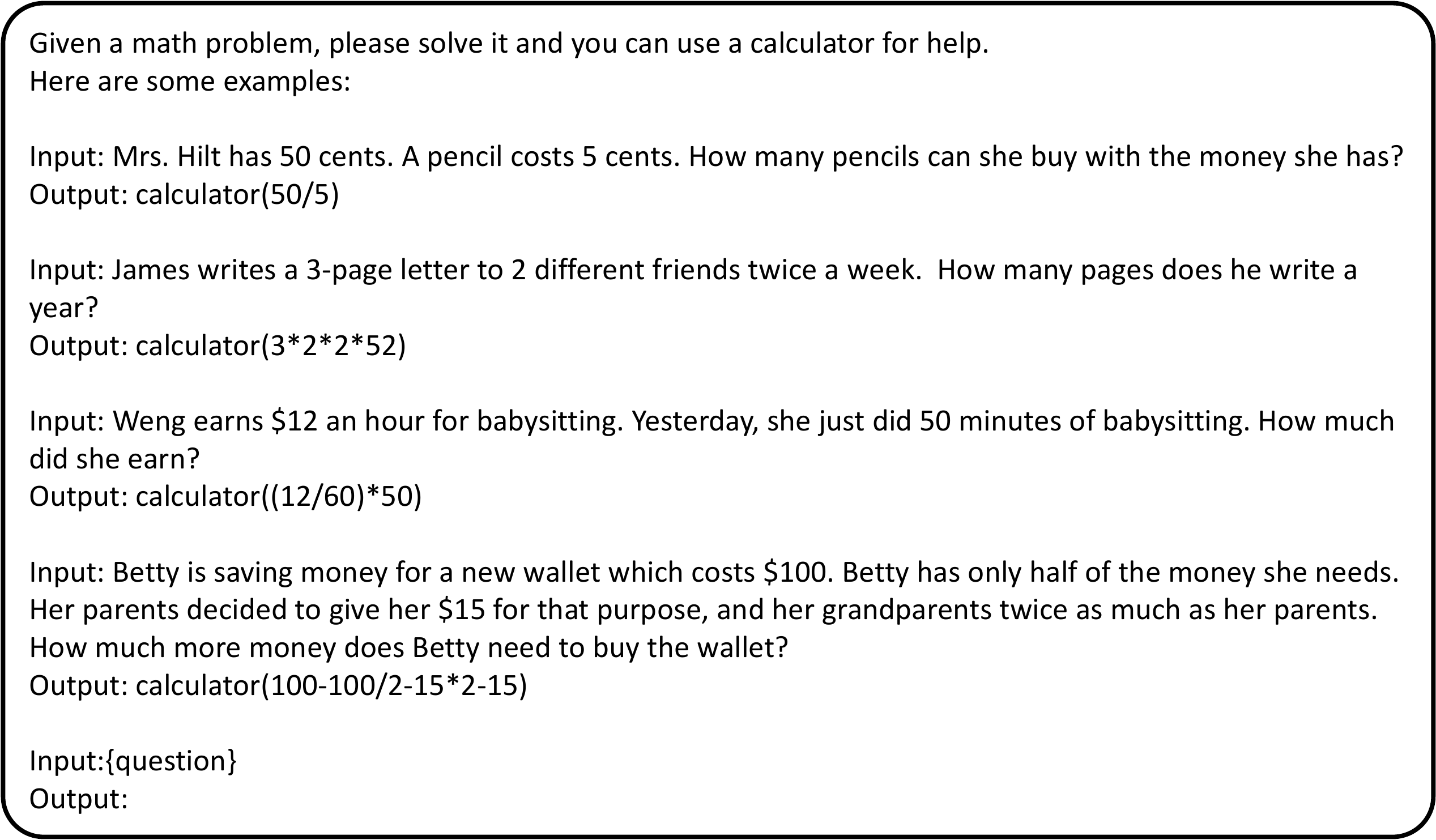}}
    \caption{Prompt used for Math Reasoning to generate candidate responses.}
    \label{fig:math_prompt_2}
\end{figure*}

\begin{figure*}[t!]
    \centering
    \resizebox{.96\textwidth}{!}{
    \includegraphics{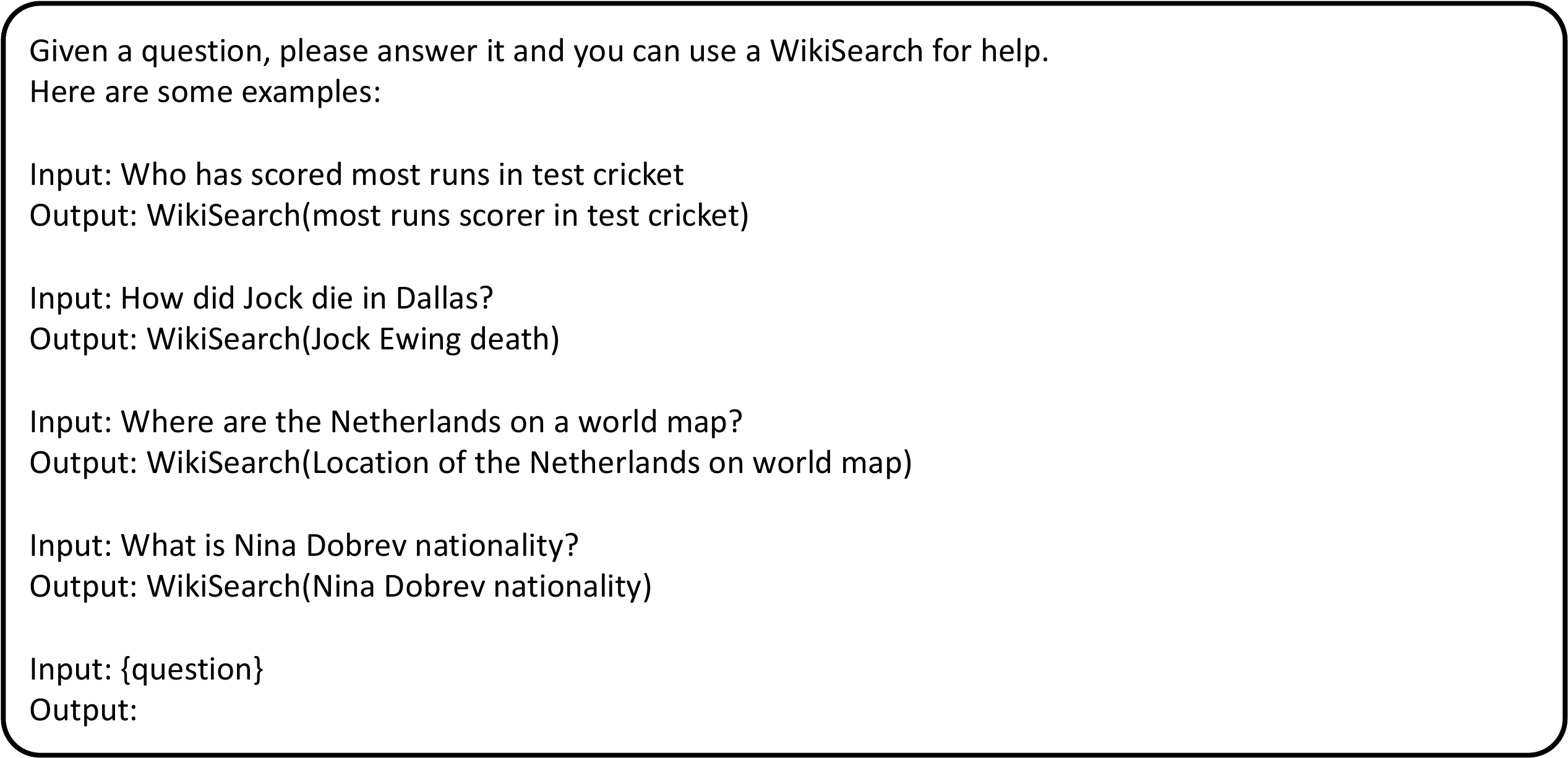}}
    \caption{Prompt used for Question Answering to generate candidate responses.}
    \label{fig:qa_prompt_2}
\end{figure*}

\begin{figure*}[t!]
    \centering
    \resizebox{.96\textwidth}{!}{
    \includegraphics{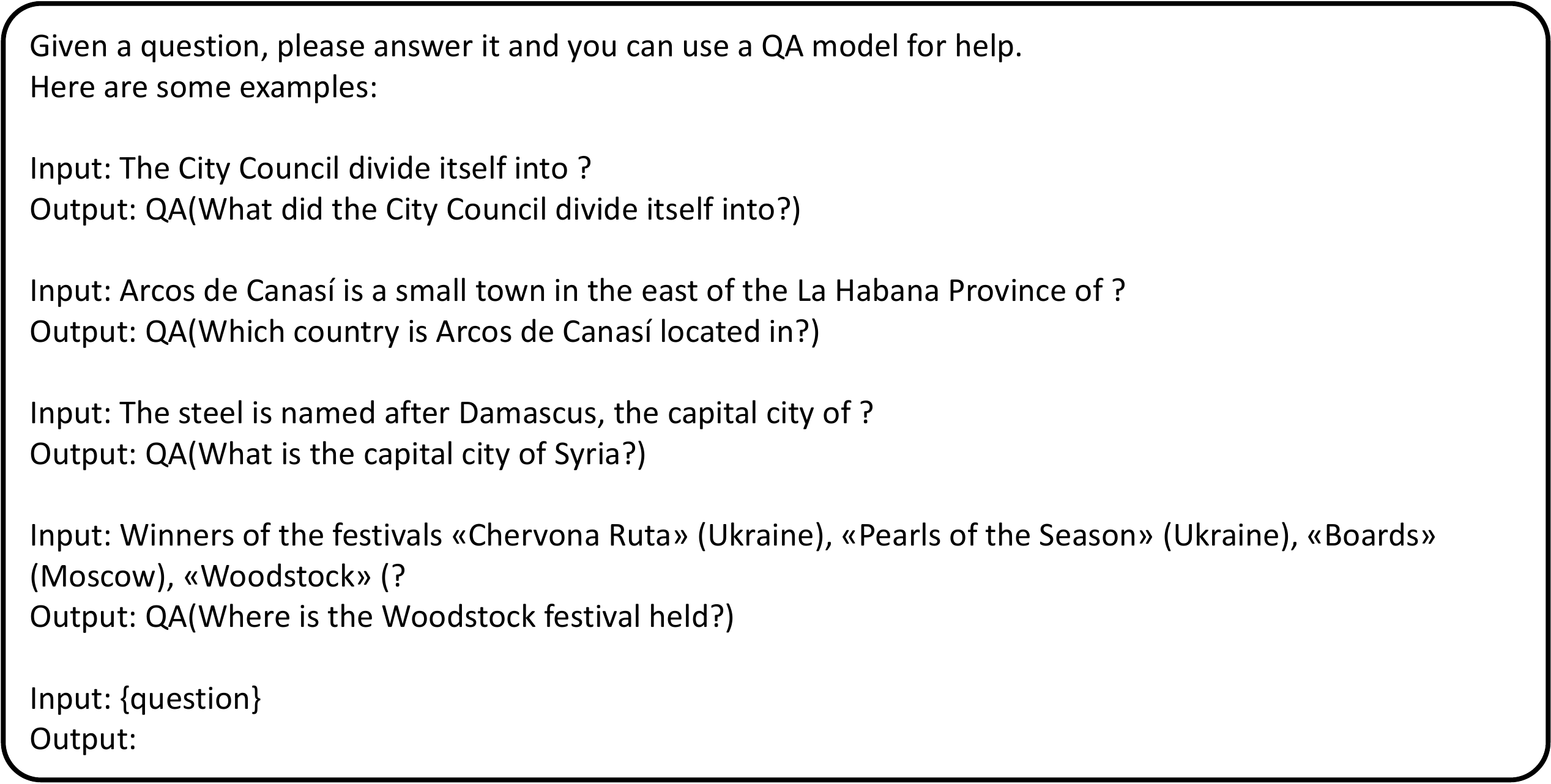}}
    \caption{Prompt used for LAMA to generate candidate responses.}
    \label{fig:lama_prompt_2}
\end{figure*}

\begin{figure*}[t!]
    \centering
    \resizebox{.96\textwidth}{!}{
    \includegraphics{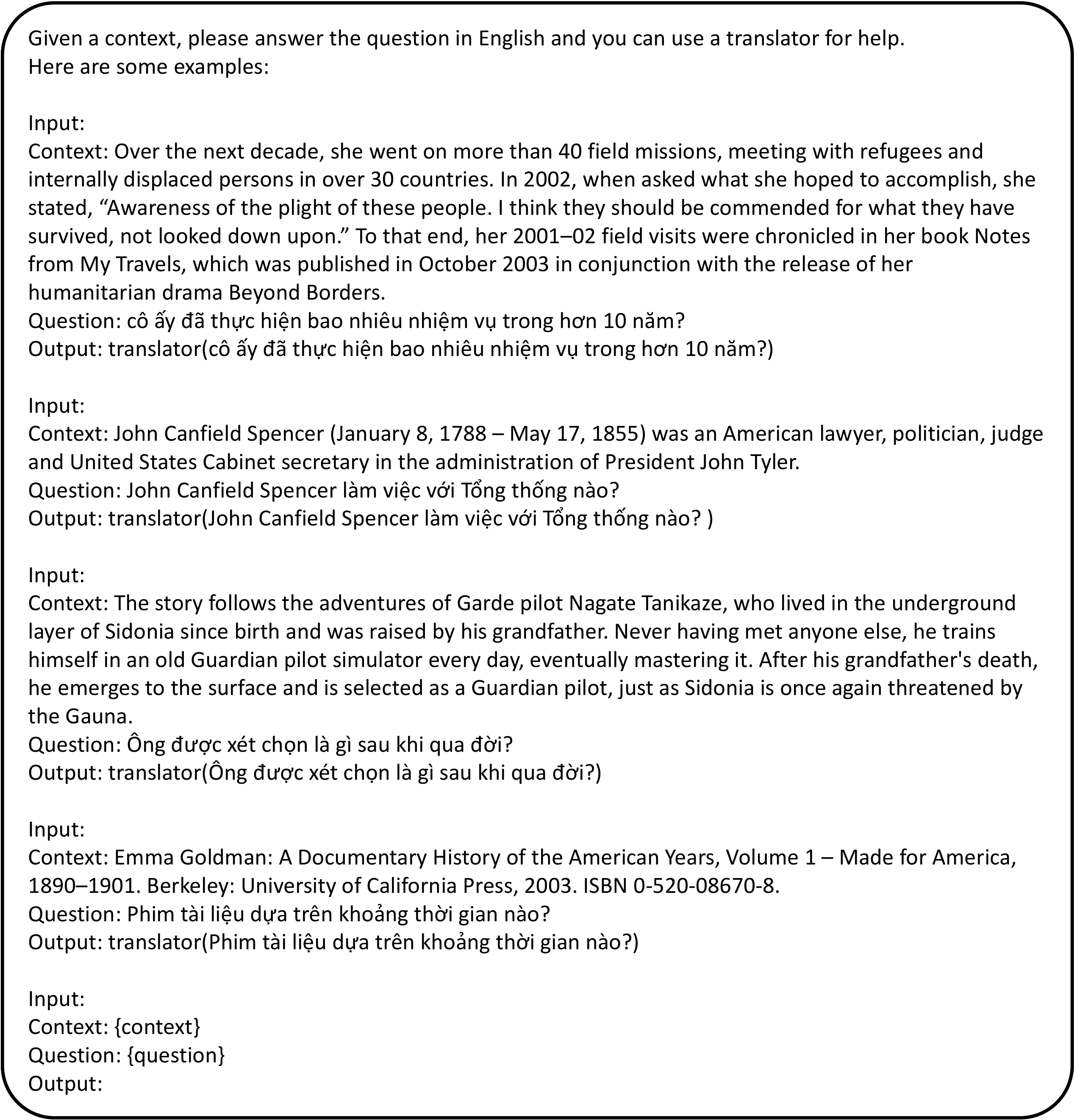}}
    \caption{Prompt used for Multilingual QA to generate candidate responses.}
    \label{fig:mlqa_prompt_2}
\end{figure*}

We show the prompt used to generate candidate responses for ChatGPT and GPT-3.5 in Figure~\ref{fig:math_prompt_2}-\ref{fig:mlqa_prompt_2}.
We use the same instructions in Figure~\ref{fig:case_study} to generate candidate responses for Vicuna and Alpaca.

\subsection{Training Details}
\label{app:training}
The hyperparameters we use to train ChatGLM-6B are shown in Table~\ref{tab:chatglm_para} and Alpaca-7B, Vicuna-7B are shown in Table~\ref{tab:alpaca_para}.
We present the training loss variations of Vicuna-7B in stages \1 and \2 of \textsc{\ours-mix} in Figure~\ref{fig:sft_loss}\&\ref{fig:rl_loss}.
During training, we observe that despite the decrease in training loss, prolonged reinforcement learning training will result in a significant performance loss.
Typically, the model achieves optimal performance within the first 10-40 steps.

\end{document}